\documentclass[lettersize,journal]{IEEEtran}
\usepackage{amsmath,amsfonts}
\usepackage{algorithmic}
\usepackage{algorithm}
\usepackage{array}
\usepackage[caption=false,font=normalsize,labelfont=sf,textfont=sf]{subfig}
\usepackage{textcomp}
\usepackage{stfloats}
\usepackage{url}
\usepackage{verbatim}
\usepackage{graphicx}
\usepackage{cite}
\usepackage{booktabs}
\usepackage{nicematrix}
\usepackage{textcomp}
\usepackage{ifthen}
\usepackage{xparse}
\usepackage[export]{adjustbox}

\newcommand{\hp}[1]{\hphantom{#1}}

\newcommand{\shadow}[1]{{\setlength{\fboxsep}{1pt}\colorbox{gray!30}{#1}}} 
\newcommand{\noshadow}[1]{{\setlength{\fboxsep}{1pt}\colorbox{white}{#1}}}
\NewDocumentCommand{\fshadow}{O{white}O{white}m}{{\setlength{\fboxsep}{1pt}\fcolorbox{#1}{#2}{#3}}}

\begin{document}

\title{How Good Is Neural Combinatorial Optimization? A Systematic Evaluation on the Traveling Salesman Problem}

\author{
	    Shengcai~Liu,
        Yu~Zhang,
        Ke~Tang,
        and~Xin~Yao \\
		Southern University of Science and Technology, China
\thanks{Corresponding author: Ke Tang (tangk3@sustech.edu.cn).}}

\markboth{Journal of \LaTeX\ Class Files,~Vol.~14, No.~8, August~2021}%
{Shell \MakeLowercase{\textit{et al.}}: A Sample Article Using IEEEtran.cls for IEEE Journals}


\maketitle

\begin{abstract}
Traditional solvers for tackling combinatorial optimization (CO) problems are usually designed by human experts.
Recently, there has been a surge of interest in utilizing deep learning, especially deep reinforcement learning, to automatically learn effective solvers for CO.
The resultant new paradigm is termed neural combinatorial optimization (NCO).
However, the advantages and disadvantages of NCO relative to other approaches have not been empirically or theoretically well studied.
This work presents a comprehensive comparative study of NCO solvers and alternative solvers.
Specifically, taking the traveling salesman problem as the testbed problem, the performance of the solvers is assessed in five aspects, i.e., effectiveness, efficiency, stability, scalability, and generalization ability.
Our results show that the solvers learned by NCO approaches, in general, still fall short of traditional solvers in nearly all these aspects.
A potential benefit of NCO solvers would be their superior time and energy efficiency for small-size problem instances when sufficient training instances are available.
Hopefully, this work would help with a better understanding of the strengths and weaknesses of NCO and provide a comprehensive evaluation protocol for further benchmarking NCO approaches in comparison to other approaches. 
\end{abstract}

\begin{IEEEkeywords}
  Neural Combinatorial Optimization, Deep Reinforcement Learning, Comparative Study, Evaluation Protocol, Traveling Salesman Problem
\end{IEEEkeywords}

\section{Introduction}
\IEEEPARstart{C}{ombinatorial} optimization (CO) concerns optimizing an objective function by selecting a solution from a finite solution set, with the latter encoding constraints on the solution space.
It has been involved in numerous real-world applications in logistics, supply chains, and energy \cite{korte2011combinatorial}.
From the perspective of computational complexity, many CO problems are NP-hard due to their discrete and nonconvex nature \cite{Karp72}.
In recent decades, methods for solving CO problems have been extensively developed and can be broadly categorized into exact and approximate/heuristic/meta-heuristic methods \cite{PuchingerR05}.
The former methods are guaranteed to optimally solve CO problems but suffer from an exponential time complexity.
In contrast, the latter methods seek to find good (but not necessarily optimal) solutions within reasonable computation time, i.e., they trade optimality for computational efficiency.

In general, most (if not all) of the above methods are manually designed.
By analyzing the structure of the CO problem of interest, domain experts would leverage the algorithmic techniques that most effectively exploit this structure (e.g., proposed in the literature) and then continuously refine these methods (e.g., introducing new algorithmic techniques).
Such a design process heavily depends on domain expertise and could be extremely expensive in terms of human time.
For example, although the well-known traveling salesman problem (TSP) \cite{gutin2006traveling} has been studied for approximately 70 years, its methods\cite{helsgaun2000effective,Helsgaun09,NagataK13,nagata2016population,TaillardH19} are still being actively and relentlessly updated.

\begin{figure*}[tbp]
  \centering
 \scalebox{0.9}{
	\subfloat[]{\includegraphics[width=0.46\linewidth]{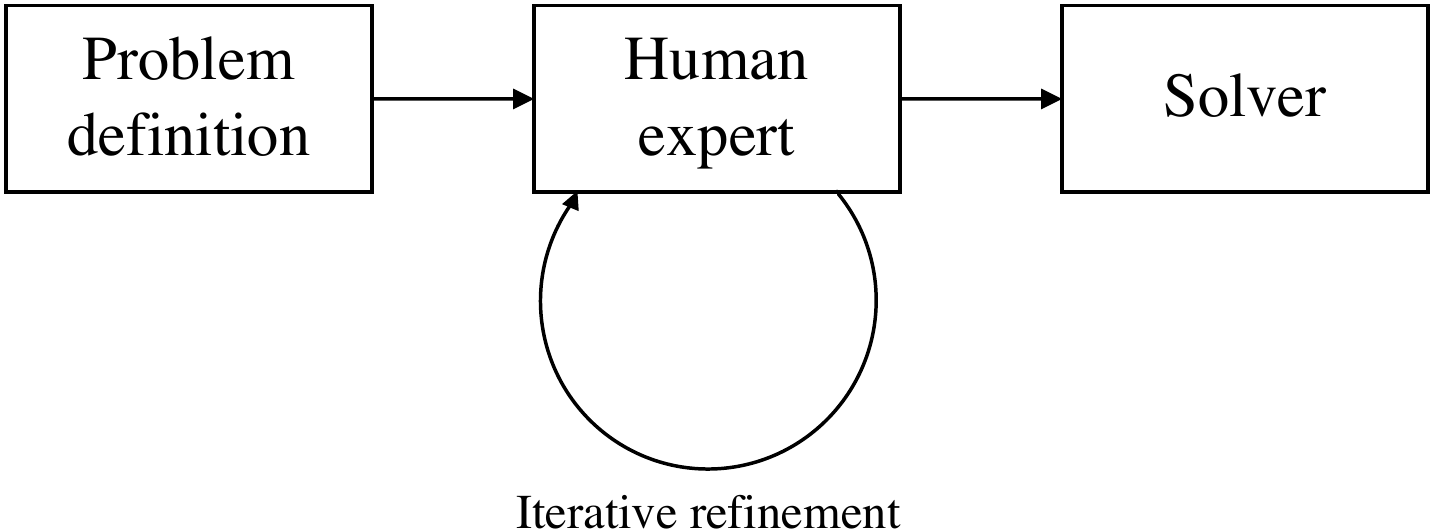}}
	\hspace{1.1cm}
  \hfil
	\subfloat[]{\includegraphics[width=0.46\linewidth]{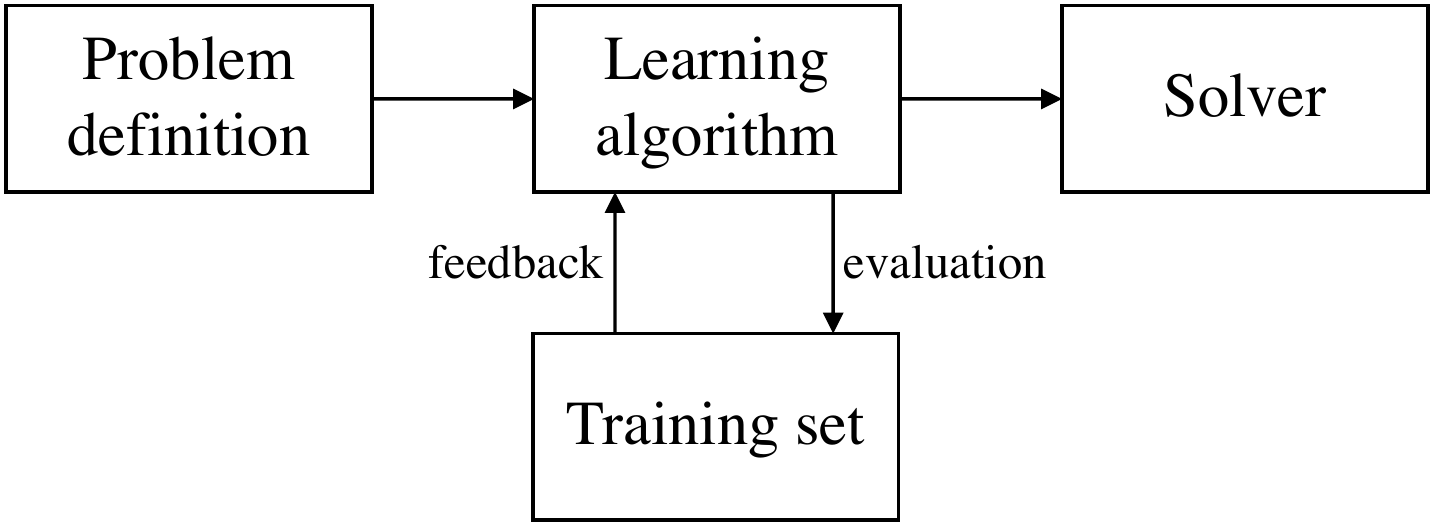}}}
	\caption{Illustrations of the two solver design paradigms. (a) Human-centered traditional paradigm. (b) Learning-centered NCO.}
	\label{fig:demonstration}
\end{figure*}

Inspired by the success of deep learning (DL) in fields such as image classification \cite{KrizhevskySH12}, machine translation \cite{BahdanauCB14}, and board games \cite{SilverSSAHGHBLB17}, recently there has been a surge of research interest in utilizing DL, especially deep reinforcement learning (DRL), to automatically learn effective methods for CO problems \cite{BengioLP21}.
The resultant new paradigm is termed neural combinatorial optimization (NCO) \cite{BelloPL0B17,KwonCYPPG21}.
For the sake of clarity, henceforth, the optimization methods (either hand-engineered or automatically learned) are called solvers and the ways to design solvers are called design approaches.
Compared to the traditional manual approach, NCO exhibits a significant paradigm shift in solver design.
As illustrated in Figure~\ref{fig:demonstration}, traditional solver design process is human-centered, while NCO is a learning-centered paradigm that develops a solver by training.
The training process of NCO essentially calibrates the parameters of the solver (model).
Although this approach induces a greater offline computational cost, the training process allows solver design to be conducted in an automated manner and thus involves much less human effort.\footnote{It is noted that NCO still requires human time and expertise to carefully construct the training set, which should sufficiently represent the target use cases of the learned solver.
However, this is not an easy task.
This point will be further discussed in Section~\ref{sec:con}.}

Despite the appealing features NCO might bring, its advantages and disadvantages relative to other approaches have not been clearly specified.
More specifically, although numerous computational experiments comparing NCO solvers with other solvers have been conducted in NCO works, they are generally non-conclusive for several reasons.
First, it is often the case that the state-of-the-art traditional solvers are missing in the comparison, which would distort the conclusion and undermine the whole validation process.
For example, the Google OR-Tools \cite{perron2019or} is widely considered by NCO works \cite{nazari2018reinforcement,KoolHW19,chen2019learning,KwonCKYGM20,ma2021learning} to be the baseline traditional solver for the vehicle routing problems (VRPs);
however, it performs far worse than the state-of-the-art solvers for VRPs \cite{AccorsiLV22}.
Second, for traditional solvers, their default configurations (parameter values) are used when comparing them with NCO solvers learned from training sets.
Such an approach neglects the fact that, when a training set is available, the performance of traditional solvers could also be significantly enhanced by tuning their parameters \cite{LiuT019,hutter2011sequential}.
In practice, it is always desirable to make full use of the available technologies to achieve the best possible performance.
In fact, with the help of the existing open-source algorithm configuration tools \cite{ansotegui2009gender,hutter2011sequential,lopez2016irace}, the tuning processes of traditional solvers can be easily automated with little human effort involved. 
Third, the benchmark instances used in the comparative studies are often quite limited in terms of problem types and sizes, making it difficult to gain insights into how these approaches would perform on problem instances with different characteristics.
For example, for TSP, the main testbed problem in NCO, most works have only reported results obtained on randomly generated instances with up to 100 nodes \cite{vinyals2015pointer,bello2016neural,deudon2018learning,KoolHW19,joshi2019efficient,KwonCKYGM20}.
In comparison, traditional TSP solvers are generally tested on problem instances collected from distinct applications, with up to tens of thousands of nodes \cite{helsgaun2000effective,Helsgaun09,NagataK13,nagata2016population,TaillardH19}.

To better understand the benefits and limitations of NCO, this work presents a more comprehensive empirical study.
Specifically, TSP is employed as the testbed problem, since it is the originally oriented problem for many widely-used architectures in NCO and thus the conclusions drawn from it could have strong implications for other problems.
Three recently developed NCO approaches and three state-of-the-art traditional TSP solvers are involved in the experiments.
These solvers are compared on five problem types with node numbers ranging from 50 to 10000.
The performance of the solvers is compared in five aspects that are critical in practice, i.e., effectiveness, efficiency, stability, scalability, and generalization ability.
In particular, the energy efficiency (in terms of electric power consumption) of the solvers is also investigated, since energy consumption is being recognized as an important factor for solver selection if the applications of solvers continue to develop.
To the best of our knowledge, this is the first comparative study of NCO approaches and traditional solvers on TSPs that
1) considers five different problem types,
2) involves problem instances with up to 10000 nodes,
3) includes tuned traditional solvers in the comparison,
and 4) investigates five different performance aspects including the energy consumption of the solvers.

The presented comprehensive empirical study has led to several interesting findings.
First, traditional solvers still significantly outperform NCO solvers in finding high-quality solutions regardless of problem types and sizes.
In particular, current NCO approaches are not adept at handling large-size problem instances and structural problem instances (e.g., clustered TSP instances).
Second, a potential benefit of NCO solvers might be their efficiency (both in time and energy).
For example, to achieve the same solution quality on small-size randomly generated problem instances, NCO solvers consume at most one-tenth of the resources consumed by traditional solvers.
Third, when the training instances cannot sufficiently represent the target cases of the  problem, both NCO solvers and tuned traditional solvers exhibit performance degradation, although the degradation is more dramatic for the former.

The remainder of the paper is organized as follows.
Section II briefly reviews the literature on NCO.
Section III explains the design of the comparative study.
Section IV presents the experimental results and analysis.
Finally, concluding remarks are given in Section V.

\section{Review of Neural Combinatorial Optimization}
Before reviewing NCO, it is useful to first quickly recap typical CO solvers.
In general, CO solvers include exact ones and approximate ones.
Typical exact solvers are based on the branch-and-bound techniques that explore the solution space by branching into sub-problems and then filtering the set of possible solutions based on the upper and lower estimated bounds of the optimal solution.
Typical approximate CO solvers are heuristics, which can be further roughly classified into constructive heuristics and improvement heuristics.
The former incrementally builds a solution to a CO problem by adding one element at a time until a complete solution is obtained.
In contrast, the latter improves upon a given solution by iteratively modifying it.
The way of modifying a given solution is called a move operator.
In recent decades, a lot of move operators have been proposed for different CO problems. 
For a comprehensive overview of CO, interested readers are referred to \cite{korte2011combinatorial}.  

It is worth noting that the applications of neural networks to solve CO problems are actually not new.
The earlier works \cite{hopfield1985neural} from the 80s in the last century focused on using Hopfield neural networks (HNNs) to solve small-size TSP instances, which were later extended to other problems \cite{smith1999neural}.
The main limitation of HNN-based approaches is that they need to use a separate network to solve each problem instance.

The term NCO refers to the series of works that utilize DL to learn a solver (model) to solve a set of different problem instances.
According to the types of the learned solvers, the existing NCO approaches can be categorized into learning constructive heuristics (LCH), learning improvement heuristics (LIH), and learning hybrid solvers (LHS).
As the names suggest, the solvers learned by LCH approaches and LIH approaches are constructive heuristics and improvement heuristics, respectively.
Compared to traditional heuristics, their main differences are that the heuristic rules are no longer manually designed but are instead automatically learned.
For example, the well-known greedy constructive heuristic for TSPs always selects the closest point for insertion, while LCH approaches learn a deep neural network (DNN) to score each point and finally select the point with the highest score for insertion.
Compared to the manually designed greedy heuristic rule, the DNN model is trained with data and unnecessarily exhibits greedy behavior.
Finally, LHS approaches seek to learn solvers that are hybrids of learning models and traditional solvers.
The following sections will introduce these NCO approaches, mainly focusing on the key works.
For a comprehensive survey of this area, interested readers are referred to \cite{BengioLP21,mazyavkina2021reinforcement}.

\subsection{Learning Constructive Heuristics}
\subsubsection{Pointer Network-based Approaches}
As the seminal work, Vinyals et al. \cite{vinyals2015pointer} introduced a sequence-to-sequence model, dubbed pointer network (Pr-Net), for solving TSPs defined on the two-dimensional plane.
Specifically, Pr-Net is composed of an encoder and a decoder, and both of them are recurrent neural networks.
Given a TSP instance, the encoder parses all the nodes in it and outputs an embedding (a real-valued vector) for each of them.
Then, the decoder repeatedly uses an attention mechanism, which has been successfully applied to machine translation \cite{BahdanauCB14}, to output a probability distribution over these previously encoded nodes, eventually obtaining a permutation over all the nodes, i.e., a solution to the input TSP instance.
This approach allows the network to be used for problem instances with different sizes.
However, Pr-Net is trained by supervised learning (SL) with precomputed near-optimal TSP solutions as labels.
This could be a limiting factor, since in real-world applications such solutions of CO problems might be difficult to obtain.
To overcome the above limitation, Bello et al. \cite{bello2016neural} proposed training Pr-Net with reinforcement learning (RL).
In their implementation, the tour length of the partial TSP solution is used as the reward signal.

Another limitation of Pr-Net is that it treats the input as a sequence, while many CO problems have no natural internal ordering.
For example, the order of the nodes in a TSP instance is actually meaningless; instead, it should be better viewed as a set of nodes.
To address this issue, Nazari et al. \cite{nazari2018reinforcement} replaced the recurrent neural network in the encoder of Pr-Net, which was supposed to capture the sequential information contained in the input, with a simple embedding layer that was invariant to the input sequence.
In addition, the authors extended Pr-Net to solve VRPs, which differ from TSPs since VRPs involve dynamically changing properties (e.g., the demands of nodes) during the solution construction process.
Specifically, the proposed model passes both the static properties (coordinates) and the dynamic properties (demands) of nodes and outputs an embedding for each of them.
Then, at each decoding step, the decoder produces a probability distribution over all nodes while masking the serviced nodes and the nodes with demands that are larger than the remaining vehicle load.

\subsubsection{Transformer-based Approaches}
In addition to the sequence-to-sequence model, the well-known transformer architecture \cite{vaswani2017attention} has also been applied to solve CO problems.
In particular, transformer also follows the encoder-decoder framework but involves a so-called multi-head attention mechanism to extract deep features from the input.
Such a mechanism has been used by Deudon et al. \cite{deudon2018learning} to encode the nodes of TSP instances.
Moreover, for sequentially decoding nodes, the authors of \cite{deudon2018learning} proposed using only the last three decoding steps (i.e., the last three selected nodes) to obtain the reference vector for the attention mechanism, thus reducing the computational complexity.
A similar model to that of \cite{deudon2018learning} was implemented by Kool et al. \cite{KoolHW19}.
Notably, the authors adjusted the model for many different CO problems including TSPs, prize collecting TSPs, VRPs, and orienteering problems, to accommodate their special characteristics.
Additionally, the authors proposed an enhanced RL training procedure that used a simple rollout baseline and exhibited superior performance over \cite{deudon2018learning,nazari2018reinforcement}.

Based on the model proposed in \cite{KoolHW19}, many subsequent works have improved it to achieve better solution quality or extended it to solve other VRP variants.
For example, Peng et al. \cite{peng2019deep} adapted the model to re-encode the nodes during the solution construction process, obtaining better solution quality than the original model.
A similar idea was implemented by Xin et al. \cite{xin2020step}, where the authors proposed changing the attention weights of the visited nodes in the encoder instead of completely re-encoding them.
Another interesting work was completed by Li et al. \cite{li2020deep}, where the authors considered the multi-objective TSPs.
They first decomposed the multi-objective problem into a series of single-objective sub-problems and then used a Pr-Net to sequentially solve each sub-problem, where the network weights were shared among neighboring sub-problems.
Finally, motivated by the fact that an optimal solution to a VRP instance, in general, has many different representations, Kwon et al. \cite{KwonCYPPG21} introduced a modified RL training procedure to force diverse rollouts toward optimal solutions.
The resultant approach, called POMO, is currently one of the strongest NCO approaches for learning constructive heuristics for TSPs and VRPs.

\subsubsection{Graph Neural Network-based Approaches}
Another line of works leverages graph neural networks (GNNs) \cite{wu2020comprehensive} to address the aforementioned issue of having an order-invariant input.
Specifically, GNNs deal with graphs as inputs without considering the order of input sequence.
Khalil et al. \cite{khalil2017learning} introduced a GNN model for solving several graph CO problems including maximum cut problems, minimum vertex cover problems, and TSPs.
The model, trained with RL, learns the embeddings of the nodes in the input problem instances, and then greedily selects nodes to construct a complete solution.
It is also possible to integrate the node embeddings learned by GNNs into Pr-Net, as shown by Ma et al. \cite{ma2019combinatorial}.
Based on \cite{khalil2017learning}, subsequent works have extended GNNs to solve many other CO problems defined on graphs.
For example, Li et al. \cite{li2018combinatorial} utilized GNNs to solve the maximal independent set problems and maximal clique problems.
Unlike TSPs, for these problems the goal is not to find a permutation of nodes but a subset of nodes.
Hence, instead of sequentially extending a solution, the authors used SL to train a graph convolutional network (GCN) to directly output an estimate of the probability of selecting each point and then utilized a guided tree search to construct a feasible solution based on these estimates.
A similar work was brought forward by Joshi et al. \cite{joshi2019efficient}, where the authors trained a GNN by SL to predict the probability of an edge being in the final TSP solution and then constructed a feasible tour by beam search.

\subsubsection{Discussion}
Due to their frameworks of sequentially encoding and decoding, Pr-Net-based approaches and transformer-based approaches are intrinsically suitable for handling permutation-based problems (e.g., TSPs and VRPs), where the orders of node selection form problem solutions.
Among these two types of approaches, transformer-based approaches can achieve better performance mainly due to their advanced multi-head attention mechanism.
In contrast, GNN-based approaches are suitable for handling CO problems defined on graphs and have no requirement regarding the sequential characteristics of the problems.

Overall, compared to LIH and LHS, LCH requires the least expert knowledge about the problems to be solved and therefore has the most potential to become a domain-independent solver design framework. 
However, with respect to obtaining high-quality solutions, current solvers learned by LCH approaches still perform worse than those learned by LIH approaches \cite{wu2021learning,ma2021learning} and LHS approaches \cite{xin2021neurolkh}.

\subsection{Learning Improvement Heuristics}
Unlike LCH approaches that learn models to sequentially extend a partial solution for a given problem instance, LIH approaches seek to learn a policy that manipulates local search operators to improve a given solution.
Nonetheless, the encoder/decoder models used by LIH approaches are still similar to those used by LCH approaches.

In an early work, Chen and Tian \cite{chen2019learning} proposed learning two models to control the 2-opt operator, which is a conventional move operator for VRPs.
Specifically, the region-picking model selects a fragment of a solution to be improved and the rule-picking model selects a rewriting rule to be applied to the region.
Both models are trained by RL and the solution is improved continuously until it converges. 

Many subsequent works have improved upon \cite{chen2019learning} to achieve better solution quality.
For example, Costa et al. \cite{pmlr-v129-costa20a} proposed learning separate embeddings for nodes and edges in the solution; Wu et al. \cite{wu2021learning} simplified the approach by learning only one model to select node pairs that are subject to the 2-opt move operator.
Another notable work is that of Lu et al. \cite{lu2019learning}.
Unlike previous approaches, the authors of \cite{lu2019learning} proposed learning a model to control several different move operators and applied a random permutation operator to the solution if the quality improvement could not reach the threshold.
Finally, motivated by the circularity and symmetry of VRP solutions (i.e., cyclic sequences), Ma et al. \cite{ma2021learning} proposed a cyclic positional encoding mechanism to learn embeddings for positional features, which are independent of the node embeddings.
The decoder and the employed move operators are similar to those of \cite{wu2021learning}.
The resultant LIH approach, called DACT, has achieved superior performance on solving VRPs in comparison with other LCH approaches and LIH approaches.

Overall, compared to LCH, LIH integrates more expert knowledge about the problems (move operators) and can achieve better solution quality than the former.
On the other hand, the application scope of LIH approaches is inevitably limited by the operators they integrate.
For example, the above-mentioned approaches cannot be applied to CO problems without sequential characteristics (e.g., maximum cut problems and minimum vertex cover problems) because their move operators are inapplicable to these problems.
Moreover, because the solvers learned by LIH approaches employ an iterative local search procedure, they need to consume more computation time than the solvers learned by LCH approaches \cite{ma2021learning}.

\subsection{Learning Hybrid Solvers}
As aforementioned, LHS approaches seek to learn solvers that are hybrids of learning models and traditional solvers.
It is worth mentioning that the integration of learning models (such as neural networks) into solvers is a long-standing research topic.
For example, many studies have been conducted on integrating HNNs into evolutionary algorithms (EAs) \cite{salcedo2008assignment,salcedo2004hybrid}.
Below the research line of using DL and DRL to train such solvers is reviewed.

One early example is the neural large neighborhood search (NLNS) of Hottung and Tierney \cite{hottung2019neural}, which integrates a learning model into the well-known large neighborhood search (LNS) algorithm.
Specifically, the use of extended/large neighborhood structures has widely proved to be effective for obtaining high-quality solutions to CO problems \cite{tang2009memetic,yao1991simulated}.
LNS \cite{shaw1997new} is a typical algorithm framework that follows this idea.
It explores the solution space by iteratively applying destroy-and-repair operators to a starting solution and has exhibited strong performance on a number of VRP variants.
NLNS uses an attention-based model trained with RL as the repair operator for LNS.
Later, Chen et al. \cite{chen2020dynamic} and Gao et al. \cite{gao2020learn} introduced two different variants of NLNS.
The former trains a hierarchical recursive GCN as the destroy operator, while the latter uses an elementwise GNN with edge embedding as the destroy operator.
Both approaches adopt a fixed repair operator that simply inserts the removed nodes into the solution according to the minimum cost principle.

In addition to LNS, another notable example is the Lin-Kernighan-Helsgaun (LKH) algorithm \cite{helsgaun2000effective}, which is widely recognized as a strong solver for TSPs.
During the solution process, LKH iteratively searches for $\lambda$-opt moves based on a small candidate edge set to improve the existing solution.
Zheng et al. \cite{zheng2021combining} proposed training a policy that helps LKH select edges from the generated candidate set. However, the policy is trained for each instance instead of a set of instances.
Later, Xin et al. \cite{xin2021neurolkh} proposed training a GNN with SL to predict edge scores, based on which LKH can create the candidate edge set and transform edge distances to guide the search process.
The resultant LHS approach, called NeuroLKH, has learned solvers that remarkably outperform the original LKH algorithm in obtaining high-quality solutions when solving TSPs and VRPs.

Compared to LCH and LIH, LHS integrates the most expert knowledge (traditional solvers) about the problems and can obtain the best solution quality \cite{xin2021neurolkh}.
However, since LHS relies on the existing solvers, its application scope is limited to the problems for which strong solvers exist.
Moreover, a LHS approach is generally specifically tailored for a solver (e.g., NeuroLKH is tailored for LKH) and it is difficult to be extended to other solvers/problems.

\section{Comparative Studies}
This section first explains the design principle and the overall framework of the comparative study, then elaborates on the details, and finally summarizes the main differences between this study and the previous ones.

Specifically, the whole study is designed to simulate two typical scenarios that arise in practice when a practitioner is faced with a CO problem to solve.
In the first scenario, one is aware of the target problem instances that the solver is expected to solve and can collect sufficient training instances to represent them.
As an illustrative example, consider a delivery company that needs to solve TSP instances for the same city on a daily basis, with only slight travel time differences across the instances due to varying traffic conditions.
In this example, one can use the accumulated instances to sufficiently represent the target use cases of the TSP solver. 
Suppose that the company expands its delivery business to another city, which differs from the first city in terms of their sizes, traffic conditions, and customer distributions.
Then, the decision maker of the company faces the second scenario, in which the information of the target use cases of the solver is unavailable; thus, the decision maker expects the solver to handle problem instances with a broad range of problem characteristics.

From the perspective of computational study, the first scenario corresponds to the setting where the training instances and the testing instances have the same problem characteristics.
NCO approaches are intrinsically appropriate for learning solvers in this scenario.
On the other hand, traditional solvers can be directly applied to the testing instances or can first be tuned with the training instances and then tested.
Furthermore, in this scenario practitioners are often concerned about the following aspects regarding the performance of the solvers.
\begin{enumerate}
  \item \textbf{Effectiveness}: the extent to which the solver can solve the problem instances, generally measured by solution quality.
  \item \textbf{Efficiency}: the computational resources (energy and computation time) consumed by the solver.
  \item \textbf{Stability}: the extent to which the output of the solver is affected by its internal randomness.
  \item \textbf{Scalability}: the problem sizes that the solver can handle.
  This is a natural performance consideration for traditional solvers. For NCO, it can be easily mixed up with generalization (see its definition below). More specifically, the scalability of an NCO approach refers to its ability to learn solvers as the problem size grows.
\end{enumerate}
Unlike the first scenario, the second scenario corresponds to the experimental setting where the testing instances significantly differ from the training instances.
In this scenario, practitioners expect the solvers to generalize well from training instances to unseen testing instances.
\begin{enumerate}
  \setcounter{enumi}{4}
  \item \textbf{Generalization}: how the learned solver would perform on instances that have different characteristics (e.g., problem sizes) from those of the training instances. 
\end{enumerate}

In the comparative study, NCO solvers and traditional solvers are evaluated in the above two scenarios.
In particular, this work takes TSP as the testbed problem to elaborate on the design of the experiments.
As a conventional CO problem, TSP has been studied for many years, and a number of strong traditional solvers have been developed for it \cite{helsgaun2000effective,Helsgaun09,NagataK13,nagata2016population,TaillardH19}.
More importantly, TSP has been the testbed problem for nearly all leading architectures in NCO \cite{vinyals2015pointer,bello2016neural,KoolHW19,KwonCKYGM20,ma2021learning,xin2020step,xin2021neurolkh};
thus, the most recently proposed NCO approaches that have achieved strong performance can be included in the experiments, and the conclusions drawn from this study may also have strong implications for other CO problems.

Since most NCO solvers are trained to handle the EUC-2D TSP instances, where the nodes are defined on a two-dimensional plane and the distances between two nodes are the same in both directions, this study also considers the EUC-2D TSP instances.

\subsection{Overall Framework}
The whole experiments aim to answer the following five research questions.
\begin{enumerate}
  \item \textbf{Q1}: In the first scenario, how would the solvers perform on small-size problem instances?
  \item \textbf{Q2}: In the first scenario, how would the solvers perform on medium/large-size problem instances?
  \item \textbf{Q3}: In the second scenario, how would the solvers generalize over different problem types (i.e., characterized by node distributions)?
  \item \textbf{Q4}: In the second scenario, how would the solvers generalize over different problem sizes?
  \item \textbf{Q5}: In the second scenario, how would the solvers generalize over different problem types and sizes?
\end{enumerate}

Specifically, the first two questions are concerned with the effectiveness, efficiency, stability, and scalability of the solvers in the first scenario where the training instances and the testing instances have the same problem characteristics.
The other three questions are concerned with the generalization ability of the solvers in the second scenario, where the training instances and the testing instances have different problem characteristics.

Each of the above questions is investigated in a separate group of experiments, denoted as Exp\_1/2/3/4/5.
Note that throughout the experiments, training instances are only used for learning NCO solvers or tuning traditional solvers, and all the solvers are tested on the testing instances.
The training/testing sets in each group of experiments, the compared methods, and the evaluation metrics are further elaborated below.

\subsection{Benchmark Instances}
Two different sources for obtaining TSP instances were considered: data generation and existing benchmark sets.
Specifically, for data generation, the \textit{portgen} generator which has been used to create testbeds for the 8-th DIMACS Implementation Challenge \cite{gutin2006traveling} and the \textit{ClusteredNetwork} generator from the \textit{netgen} R-package \cite{bossek2015netgen} were used.
\begin{enumerate}
  \item The \textit{portgen} generator generates a TSP instance (called a \textit{rue} instance) by uniformly and randomly placing points on a two-dimensional plane.
  \item The \textit{ClusteredNetwork} generator generates a TSP instance (called a \textit{clu} instance) by randomly placing points around different central points.
\end{enumerate}
Three  benchmark sets, i.e., \textit{TSPlib}, \textit{VLSI}, and \textit{National}, were used:
\begin{enumerate}
  \item \textit{TSPlib} \cite{reinelt1991tsplib}: a widely used benchmark set of instances drawn from industrial applications and geographic problems featuring the locations of cities (nodes) on maps.
  \item \textit{VLSI}: a benchmark set of instances extracted from the very-large-scale integration design data of the Bonn Institute.
  \item \textit{National}: a benchmark set of instances extracted from the maps of different countries.\footnote{All three benchmark sets are available at \url{http://www.math.uwaterloo.ca/tsp/data.}}
\end{enumerate}
For all the generated instances, Concorde \cite{applegate2006concorde}, an exact TSP solver, was used to obtain their optimal solutions.\footnote{Concorde is available at \url{https://www.math.uwaterloo.ca/tsp/concorde.html}.}
For the instances belonging to the existing benchmark sets, their optimal solutions or best-known solutions (in case the optimal solutions are unknown) were collected and used.

\begin{table*}[tbp]
  \caption{The training sets and the testing sets (separated by ``$\vert$'') in five groups of experiments.
	``$\mathrm{Exist.\ Bench.}$'' refers to the testing set containing 30 instances selected from the existing benchmark sets.}
  \centering
  \scalebox{0.9}{
  \begin{NiceTabular}{c|p{0.13\textwidth}p{0.13\textwidth}|p{0.105\textwidth}p{0.105\textwidth} p{0.105\textwidth}}
  \toprule
  \Block{2-1}{Experiment Group} & \Block[c]{1-2}{Scenario 1} & &  \Block[c]{1-3}{Scenario 2} & & \\
  \cmidrule(rl){2-3}
  \cmidrule(rl){4-6}
  & \Block{1-1}{Exp\_1} & \Block{1-1}{Exp\_2} & \Block{1-1}{Exp\_3} & \Block{1-1}{Exp\_4} & \Block{1-1}{Exp\_5} \\
  \midrule
  \Block{4-1}{Training Set $\vert$ Testing Set} & \Block{1-1}{\textit{rue}-50 $|$ \textit{rue}-50} & \Block{1-1}{\textit{rue}-500 $|$ \textit{rue}-500} & \Block{1-1}{\textit{rue}-100, \textit{mix}-100 $|$ \textit{clu}-100} & \Block{4-1}{\textit{rue}-50 $|$ \textit{rue}-100 \\ \textit{clu}-50 $|$ \textit{clu}-100} & \Block{4-1}{\textit{rue}-1000 $|$ Exist. Bench. \\ \textit{clu}-1000 $|$ Exist. Bench. \\ \textit{mix}-1000 $|$ Exist. Bench.} \\
   & \Block{1-1}{\textit{clu}-50 $|$ \textit{clu}-50} & \Block{1-1}{\textit{clu}-500 $|$ \textit{clu}-500} & \Block{1-1}{\textit{clu}-100, \textit{mix}-100 $|$ \textit{rue}-100} & &  \\
   & \Block{1-1}{\textit{rue}-100 $|$ \textit{rue}-100} & \Block{1-1}{\textit{rue}-1000 $|$ \textit{rue}-1000} & \Block{1-1}{\textit{rue}-1000, \textit{mix}-1000 $|$ \textit{clu}-1000} &  &  \\
   & \Block{1-1}{\textit{clu}-100 $|$ \textit{clu}-100} & \Block{1-1}{\textit{clu}-1000 $|$ \textit{clu}-1000} & \Block{1-1}{\textit{clu}-1000, \textit{mix}-1000 $|$ \textit{rue}-1000} & & \\
   \midrule
    Description & \Block{1-2}{The training instances and the testing instances \\ have the
    same problem characteristics}
    & & \Block{1-3}{The training instances and the testing instances differ in
    either \\problem types, problem sizes, or both.} & &\\
    \bottomrule
    \end{NiceTabular}}
  \label{tab:benchmark_sets}
\end{table*}

Based on the above data generation/collection procedure, the training/testing sets in each of the five groups of experiments were constructed as follows (also summarized in Table~\ref{tab:benchmark_sets}):
\begin{enumerate}
  \item \textbf{Exp\_1}: Two problem sizes (50 and 100) and two problem types (\textit{rue} and \textit{clu}) were considered.
  Consequently, four combinations were produced, denoted as \textit{rue}/\textit{clu}-50/100.
  For each of them, following the common practice in NCO \cite{KoolHW19,KwonCKYGM20,ma2021learning}, one million training instances and 10000 testing instances were generated.
  For the \textit{clu} instances, the number of clusters was set to $n/10$, where $n$ was the problem size.
  \item \textbf{Exp\_2}: The whole procedure for constructing training/testing sets was exactly the same as that used in Exp\_1, except that the considered problem sizes were 500/1000 and the testing set size was 1000.
  Besides, for the \textit{clu} instances, the number of clusters was set to $n/100$.
  \item \textbf{Exp\_3}: Unlike Exp\_1 and Exp\_2, in this experiment, the testing instances differed from the training instances in problem types.
  Specifically, two problem sizes (100 and 1000)  and two problem types (\textit{rue} and \textit{clu})  were considered.
  The solvers learned on the training set of \textit{rue}-100/1000 instances would be tested on the testing set of \textit{clu}-100/1000 instances, and vice versa.
  Moreover, in addition to the \textit{rue} and \textit{clu} training sets, another training set called \textit{mix} was also used, which contained half \textit{rue} instances and half \textit{clu} instances.
  \item \textbf{Exp\_4}: In this experiment, the testing instances differed from the training instances in problem sizes.
  Specifically, two problem sizes (50 and 100) and two problem types (\textit{rue} and \textit{clu}) were considered.
  The solvers learned on the \textit{rue}-50 training instances and the \textit{clu}-50 training instances would be tested on the \textit{rue}-100 testing instances and the \textit{clu}-100 testing instances, respectively.
  \item \textbf{Exp\_5}: In this experiment, problem instances selected from the \textit{TSPlib}, \textit{VLSI}, and \textit{National}  benchmark sets were used as the testing instances.
  Specifically, 10 instances were selected from each of these three sets, with problem sizes distributed between 1000 and 10000.
  In addition, three training sets were considered in this experiment, i.e., \textit{rue}-1000, \textit{clu}-1000, and \textit{mix}-1000.
\end{enumerate}

\begin{table}[tbp]
  \caption{The competitors in the experiments.}
  \centering
  \scalebox{1.0}{
  \begin{NiceTabular}{ll}
  \toprule
    Method & Type\\
  \midrule
    POMO~\cite{KwonCKYGM20} & Learning Constructive Heuristics (LCH)\\
    DACT~\cite{ma2021learning} & Learning Improvement Heuristics (LIH)\\
    NeuroLKH~\cite{xin2021neurolkh} & Learning Hybrid Solvers (LHS)\\
    LKH~\cite{helsgaun2000effective} &  Traditional Solver \\
    EAX~\cite{NagataK13} & Traditional Solver \\
    MAOS~\cite{XieL09} & Traditional Solver \\
    LKH (tuned) & Tuned Traditional Solver \\
    \bottomrule
    \end{NiceTabular}}
  \label{tab:compared_methods}
\end{table}

\subsection{Compared Methods}
Table~\ref{tab:compared_methods} lists all the competitors in the experiments.
For each of the three types of NCO approaches, a recently proposed approach that has achieved strong performance was considered.
Specifically, POMO \cite{KwonCKYGM20}, DACT \cite{ma2021learning}, and NeuroLKH  \cite{xin2021neurolkh} were the considered approaches for LCH, LIH, and LHS, respectively.
 According to the results reported in \cite{KwonCKYGM20} and \cite{ma2021learning}, POMO and DACT could achieve their best performance with an extra instance augmentation mechanism; thus, these variants of POMO and DACT were also considered in the experiments.
All the hyper-parameters of these approaches were set as reported in  their original papers, except that in the experiments, their batch sizes  were always tuned to fully utilize the GPU memory.

Regarding traditional solvers, except for the widely adopted LKH (version 3.0) \cite{helsgaun2000effective} in NCO works, this study included two other meta-heuristic solvers, EAX \cite{NagataK13} and MAOS~\cite{XieL09}, in the experiments.\footnote{Concorde was not included in the comparison because it needs to run for prohibitively long periods of time to solve those very-large-size problem instances (e.g., larger than 5000).
Besides, LKH, EAX, and MAOS could achieve solution quality very close to  that of Concorde, while consuming much less computation time than the latter.}
EAX is a genetic algorithm equipped with a powerful edge assembly crossover.
It has proved to outperform LKH in solving a broad range of TSP instances \cite{NagataK13}.
MAOS is a strong swarm intelligence-based TSP solver that does not contain any explicit local search heuristic.
The parameters of LKH, EAX, and MAOS were kept as  their default values in the experiments.
Moreover, LKH poses  many parameters whose values may significantly affect its performance; it is thus possible to tune these parameters on a training set to achieve better performance.
Hence, in the experiments, the tuned variants of LKH obtained by using the general-purpose automatic algorithm configuration tool SMAC \cite{hutter2011sequential}  were also considered.
Generally, the computation time needed by SMAC to tune LKH was much shorter than that needed by NCO approaches to train  their solvers.

\subsection{Evaluation Metrics}
The testing results of the solvers are reported in terms of three metrics, i.e., optimum gap, computation time, and energy.
For all three metrics, the smaller the results are, the better. Specifically, the optimum gap is defined as 
\begin{equation*}
	(Q-Q^*)/Q^*,
\end{equation*}
where $Q$ is the length of the tour found by the solver and $Q^*$ is the optimal tour length.

The computation time of a solver is the time it takes to solve all the instances in the testing set.
Note that NCO solvers would naturally benefit from running on massively parallel hardware architectures, i.e., GPUs, while in previous comparative studies \cite{bello2016neural,nazari2018reinforcement,KoolHW19,KwonCKYGM20,ma2021learning} traditional solvers were generally run on CPUs using a single thread.
To conduct fair comparisons, in the experiments, the traditional solvers and their tuned variants were also run on $k$ CPU threads to solve $k$ problem instances in parallel ($k=32$ on our reference machine).
Nevertheless, it is noted that the different programming languages adopted by NCO solvers and traditional solvers would also affect their runtime, and this cannot be avoided in our experiments.
Specifically, NCO solvers are usually implemented with Python that mixes inefficient interpreted code with efficient DL libraries (e.g., those for utilizing GPUs).
On the other hand, traditional solvers are typically implemented in highly efficient languages such as C/C++/java.
Currently, how to avoid the influence of different programming languages when comparing NCO solvers and traditional solvers is still an open question.

Finally, the energy is the electric power consumed by  a solver for solving all the instances in the testing set, which is a particularly useful metric in resource-limited cases such as embedded devices.
In the experiments, the open-source PowerJoular tool was used to record the electric power consumed by the solvers.\footnote{PowerJoular is available at \url{https://github.com/joular/powerjoular}.}

When testing the solvers, to prevent them from running for prohibitively long periods of time, the maximum runtime for solving a testing instance was set to 3600 seconds.
If a solver consumed its time budget, it would be terminated immediately and the best solution found by it would be returned.
Note that some tested solvers (e.g., EAX and LKH) involve randomized components.
In the experiments, these solvers were applied on each testing instance for 10 runs.
Then, the mean value, as well as the standard deviation of the optimum gaps over the 10 runs were recorded, which were further averaged over all the testing instances to obtain the average optimum gap and the average standard deviation on the whole testing set.

All the experiments were conducted on a server with an Intel Xeon Gold 6240 CPU (2.60 GHz, 24.75 MB of Cache) and an NVIDIA TITAN RTX GPU (24 GB of video memory) with 377 GB of RAM, running Ubuntu 18.04.
The complete experimental results, benchmark instances, NCO solvers, traditional solvers and their tuned variants, and codes for training/tuning solvers are available at \url{https://github.com/yzhang-gh/benchmarking-tsp}.

\subsection{Main Differences from Previous Comparative Studies}
In general, the above established experimental protocol could be used as a standard protocol for benchmarking NCO approaches.
More specifically, our comparative study differs from the studies presented in previous NCO works\cite{bello2016neural,nazari2018reinforcement,KoolHW19,KwonCKYGM20,ma2021learning,xin2021neurolkh} in the following aspects.
\begin{enumerate}
  \item Regarding benchmark instances, in the NCO literature, it is common to use the \textit{rue} type of instances as both training and testing instances, and some studies used \textit{TSPlib} to assess the generalization ability of their learned solvers.
  Compared to them, this study used three more types of problem instances (\textit{clu}, \textit{National}, and \textit{VLSI}) in the experiments, leading to testbed problems with much more diverse characteristics.
  Moreover, the considered problem sizes ranged from 50 to 10000, which were much larger than those in the previous NCO works.
  \item Regarding traditional solvers, in addition to the LKH solver widely adopted by NCO works, this study included two other strong solvers (EAX and MAOS) in the comparison to fully represent the state-of-the-art TSP solvers.
  To assess the potential of traditional solvers, this study also considered tuning their parameters, which to the best of our knowledge has never been considered by the existing NCO works.
  \item This study investigated five different performance aspects and introduced a new efficiency metric, i.e., electric power consumption, which could be particularly useful in energy-limited environments.
  Besides, to conduct fair comparisons in terms of time efficiency, all the NCO solvers and traditional solvers (and their tuned variants) were tested in the parallel mode to make full use of our reference machines.
  In comparison, previous comparative studies often tested traditional solvers on CPUs using a single thread. 
\end{enumerate}

\section{Experimental Results and Analysis}
This section first presents the main findings drawn from the experiments and then analyzes the results of each group of experiments in detail.

\begin{table*}[tbp]
	\caption{Testing results of Exp\_1/2.
	For each metric, the best performance is indicated in gray.
	LKH* and EAX* refer to the variants of LKH and EAX, respectively, which were terminated once they achieved the same solution quality as that of POMO solver.}
	\label{tab:exp_1}
	\centering
  \scalebox{1.1}{
	\begin{NiceTabular}{lcrrcrr}
		\toprule
		\Block{1-7}{\textbf{Exp\_1: Small-Size Problem Instances}} & & & & & &\\
		\midrule
		\Block[c]{2-1}{Method}         &
		\Block{1-3}{\textit{rue}-50}            &                       &                         &
		\Block{1-3}{\textit{clu}-50}          &                       &                         \\
		
		\cmidrule(rl){2-4}
		\cmidrule(rl){5-7}
		&
		Gap (\%) \(\pm\) std (\%)      & \Block[c]{}{Time (s)} & \Block[c]{}{Energy (J)} &
		Gap (\%) \(\pm\) std (\%)      & \Block[c]{}{Time (s)} & \Block[c]{}{Energy (J)}  \\
		\midrule
		
		POMO, no aug.                  &
		0.1185 \(\pm\) 0.0000          & \shadow{2.57}         & \shadow{290.83}         & 
		0.1353 \(\pm\) 0.0000          & \shadow{2.58}         & \shadow{292.14}          \\ 
		
		POMO, \(\times 8\) aug.        &
		0.0228 \(\pm\) 0.0000          & \noshadow{16.98}      & \noshadow{3361.87}      &
		0.0213 \(\pm\) 0.0000          & \noshadow{17.04}      & \noshadow{4193.39}      \\
		
		DACT                  &
		0.0167 \(\pm\) 0.0291          & \noshadow{1991.49}    & \noshadow{402635.40}    &
		0.1770 \(\pm\) 0.1117          & \noshadow{1921.99}    & \noshadow{393994.52}    \\
		
		DACT, \(\times 4\) aug.  &
		0.0006 \(\pm\) 0.0013          & \noshadow{8534.42}    & \noshadow{1742933.29}   &
		0.0576 \(\pm\) 0.0390          & \noshadow{8735.50}    & \noshadow{1676140.80}   \\
		
		NeuroLKH                       &
		0.0003 \(\pm\) 0.0003          & \noshadow{34.22}      & \noshadow{4006.11}      &
		0.0004 \(\pm\) 0.0007          & \noshadow{131.92}     & \noshadow{12698.78}     \\
		
		MAOS                       &
		\shadow{0.0000 \(\pm\) 0.0000}          & \noshadow{357.33}      & \noshadow{37219.44}      &
		\shadow{0.0000 \(\pm\) 0.0000}         & \noshadow{350.84}     & \noshadow{36387.72}    \\
		
		LKH                            &
		0.0035 \(\pm\) 0.0035          & \noshadow{291.22}     & \noshadow{10458.70}     &
		0.0022 \(\pm\) 0.0018          & \noshadow{257.95}     & \noshadow{9512.24}      \\
		
		EAX                            &
		\shadow{0.0000 \(\pm\) 0.0000} & \noshadow{343.79}     & \noshadow{15089.39}     &
		\shadow{0.0000 \(\pm\) 0.0000} & \noshadow{321.06}     & \noshadow{12541.47}     \\
		
		LKH*                            &
		\Block[c]{2-1}{Same as POMO}          & \noshadow{259.17}     & \noshadow{9551.16}     &
		\Block[c]{2-1}{Same as POMO}          & \noshadow{207.11}     & \noshadow{6228.23}      \\
		
		EAX*                            &
		& \noshadow{273.95}     & \noshadow{10248.02}     &
		& \noshadow{267.95}     & \noshadow{10165.34}      \\
		
		\midrule
		\Block[c]{2-1}{Method}         &
		\Block{1-3}{\textit{rue}-100}           &                       &                         &
		\Block{1-3}{\textit{clu}-100}         &                       &                           \\
		\cmidrule(rl){2-4}
		\cmidrule(rl){5-7}
		&
		Gap (\%) \(\pm\) std (\%)      & \Block[c]{}{Time (s)} & \Block[c]{}{Energy (J)} &
		Gap (\%) \(\pm\) std (\%)      & \Block[c]{}{Time (s)} & \Block[c]{}{Energy (J)}   \\
		\midrule
		POMO, no aug.                  &
		0.3646 \(\pm\) 0.0000          & \shadow{12.59}        & \shadow{2588.90}        & 
		0.4318 \(\pm\) 0.0000          & \shadow{12.83}        & \shadow{2675.18}          \\ 
		
		POMO, \(\times 8\) aug.        &
		0.1278 \(\pm\) 0.0000          & \noshadow{87.73}      & \noshadow{25873.38}     &
		0.1405 \(\pm\) 0.0000          & \noshadow{93.08}      & \noshadow{27299.18}       \\
		
		DACT                  &
		0.6596 \(\pm\) 0.5216          & \noshadow{6141.72}    & \noshadow{1269009.13}   &
		1.2220 \(\pm\) 0.4773          & \noshadow{6517.2110}  & \noshadow{1390740.24}     \\
		
		DACT, \(\times 4\) aug.  &
		\Block{1-3}{-}          &                       &                         &
		\Block{1-3}{-}          &                       &                           \\
		
		NeuroLKH                       &
		0.0004 \(\pm\) 0.0005          & \noshadow{74.90}      & \noshadow{9819.77}      &
		0.0021 \(\pm\) 0.0031          & \noshadow{309.75}     & \noshadow{29397.05}       \\
		
		MAOS                       &
		\shadow{0.0000 \(\pm\) 0.0000}          & \noshadow{451.94}      & \noshadow{55214.54}      &
		\shadow{0.0000 \(\pm\) 0.0000}          & \noshadow{447.51}     & \noshadow{54716.25}     \\

		LKH                            &
		0.0044 \(\pm\) 0.0048          & \noshadow{313.73}     & \noshadow{12868.38}     &
		0.0048 \(\pm\) 0.0040          & \noshadow{340.58}     & \noshadow{14264.64}       \\
		
		EAX                            &
		\shadow{0.0000 \(\pm\) 0.0000} & \noshadow{598.34}     & \noshadow{35145.37}     &
		\shadow{0.0000 \(\pm\) 0.0000} & \noshadow{561.41}     & \noshadow{27893.49}       \\
		
		LKH*                            &
		\Block[c]{2-1}{Same as POMO}          & \noshadow{262.99}     & \noshadow{10707.67}     &
		\Block[c]{2-1}{Same as POMO}          & \noshadow{294.30}     & \noshadow{12140.67}      \\
		
		EAX*                            &
		& \noshadow{366.99}     & \noshadow{22638.50}     &
		& \noshadow{340.67}     & \noshadow{13422.11}      \\
		
		\midrule
		\Block{1-7}{\textbf{Exp\_2: Medium/Large-Size Problem Instances}} & & & & & &\\
		\midrule
		\Block[c]{2-1}{Method}         &
		\Block{1-3}{\textit{rue}-500}            &                       &                         &
		\Block{1-3}{\textit{clu}-500}          &                       &                         \\
		\cmidrule(rl){2-4}
		\cmidrule(rl){5-7}
		&
		Gap (\textpertenthousand) \(\pm\) std (\textpertenthousand)      & \Block[c]{}{Time (s)} & \Block[c]{}{Energy (J)} &
		Gap (\textpertenthousand) \(\pm\) std (\textpertenthousand)      & \Block[c]{}{Time (s)} & \Block[c]{}{Energy (J)}  \\
		\midrule
		
		NeuroLKH                         &
		\noshadow{0.0273 \(\pm\) 0.0441} & \noshadow{242.71}     & \noshadow{21455.69}     &
		\noshadow{1.8080 \(\pm\) 2.5469} & \noshadow{2695.06}    & \noshadow{171826.14}        \\ 
		
		MAOS                       &
		0.0870 \(\pm\) 0.1591          & \shadow{133.08}      & \shadow{13720.29}      &
		0.1502 \(\pm\) 0.1674          & \shadow{111.20}     & \shadow{11728.91}     \\

		LKH                              &
		\noshadow{0.4356 \(\pm\) 0.5084} & \noshadow{143.36}       & \noshadow{17077.63}       &
		\noshadow{4.2779 \(\pm\) 3.6307} & \noshadow{313.80}     & \noshadow{35213.13}       \\
		
		LKH (tuned)                      &
		\shadow{0.0086 \(\pm\) 0.0165}   & \noshadow{162.63}     & \noshadow{19087.78}     &
		\noshadow{0.0345 \(\pm\) 0.0398} & \noshadow{279.80}     & \noshadow{32685.72}    \\
		
		EAX                              &
		\noshadow{0.0140 \(\pm\) 0.0291} & \noshadow{269.41}     & \noshadow{32370.64}     &
		\shadow{0.0006 \(\pm\) 0.0012}   & \noshadow{209.09}       & \noshadow{24456.69}   \\
		\midrule
		\Block[c]{2-1}{Method}         &
		\Block{1-3}{\textit{rue}-1000}           &                       &                         &
		\Block{1-3}{\textit{clu}-1000}         &                       &                           \\
		\cmidrule(rl){2-4}
		\cmidrule(rl){5-7}
		&
		Gap (\textpertenthousand) \(\pm\) std (\textpertenthousand)      & \Block[c]{}{Time (s)} & \Block[c]{}{Energy (J)} &
		Gap (\textpertenthousand) \(\pm\) std (\textpertenthousand)      & \Block[c]{}{Time (s)} & \Block[c]{}{Energy (J)}  \\
		\midrule
		NeuroLKH                  &
		\noshadow{0.0417 \(\pm\) 0.0510} & \noshadow{728.41}     & \noshadow{58569.96}     &
		\noshadow{1.8599 \(\pm\) 2.2236} & \noshadow{5094.77}    & \noshadow{334771.95}      \\ 
		
		MAOS                       &
		0.1263 \(\pm\) 0.1561         & \shadow{380.32}      & \shadow{37621.01}      &
		0.2096 \(\pm\) 0.1863          & \shadow{272.97}     & \shadow{27621.76}     \\
		
		LKH        &
		\noshadow{0.3620 \(\pm\) 0.3624} & \noshadow{413.43}     & \noshadow{48867.49}       &
		\noshadow{1.8761 \(\pm\) 1.2798} & \noshadow{859.39}     & \noshadow{100514.45}      \\
		
		LKH (tuned)                  &
		\noshadow{0.1732 \(\pm\) 0.1770} & \noshadow{406.16}       & \noshadow{52252.97}     &
		\noshadow{0.0460 \(\pm\) 0.0537} & \noshadow{522.41}       & \noshadow{66511.40}         \\
		
		EAX  &
		\shadow{0.0182 \(\pm\) 0.0242}   & \noshadow{620.30}     & \noshadow{75168.05}     &
		\shadow{0.0086 \(\pm\) 0.0122}   & \noshadow{631.68}     & \noshadow{75222.30}       \\
		\bottomrule
	\end{NiceTabular}}
\end{table*}

\subsection{Main Findings}
Overall, four main findings can be obtained based on the experimental results.

First, for all the TSP problem sizes and types considered in the experiments, traditional solvers still significantly outperformed NCO solvers in finding high-quality solutions (Sections~\ref{sec_exp1},~\ref{sec_exp2}, and~\ref{sec_exp3}).
Among NCO solvers, the hybrid solvers trained by LHS approaches could find much better solutions than the learned constructive and improvement heuristics.
In other words, the more expert knowledge that was integrated in an NCO solver, the better it solved the problems.
Hence, it appears that the research status in this area has not yet reached those in domains such as vision, speech, and natural language processing, where DL can learn strong models from scratch.

Second, due to their simple solving strategy (i.e., sequentially constructing a solution) and massively parallel computing mode, a major potential benefit of NCO solvers (i.e., the constructive heuristics learned by LCH approaches) is their superior efficiency (in terms of both time and energy).
In particular, on small-size randomly generated problem instances, the computational resources consumed by the learned heuristics were usually at most one-tenth of the resources consumed by traditional solvers (Section~\ref{sec_exp1}), where the latter were terminated once they achieved the same solution quality as that of the former.

Third, current LCH and LIH approaches are not suitable for handling large-size problem instances (Section~\ref{sec_exp2}) and structural problem instances (Sections~\ref{sec_exp1} and~\ref{sec_exp2}), e.g., the \textit{clu} type of TSP instances.

Fourth, parameter tuning can significantly boost the performance of traditional solvers in terms of solution quality while maintaining efficiency (Section~\ref{sec_exp3}).
However, when the training instances had different problem characteristics (problem types and sizes) from those of the testing instances, both NCO solvers and tuned traditional solvers exhibited performance degradation (Sections~\ref{sec_exp3} and~\ref{sec_exp4}), and NCO solvers suffered from far more severe performance degradation.

\begin{figure*}[tbp]
	\centering
	\resizebox{\textwidth}{!}{
		\begin{NiceTabular}{cc}
			\Block[]{}{\rotatebox{90}{\large{}Optimum gap}} & \Block{}{\includegraphics[trim=6pt 10pt 0 0,clip]{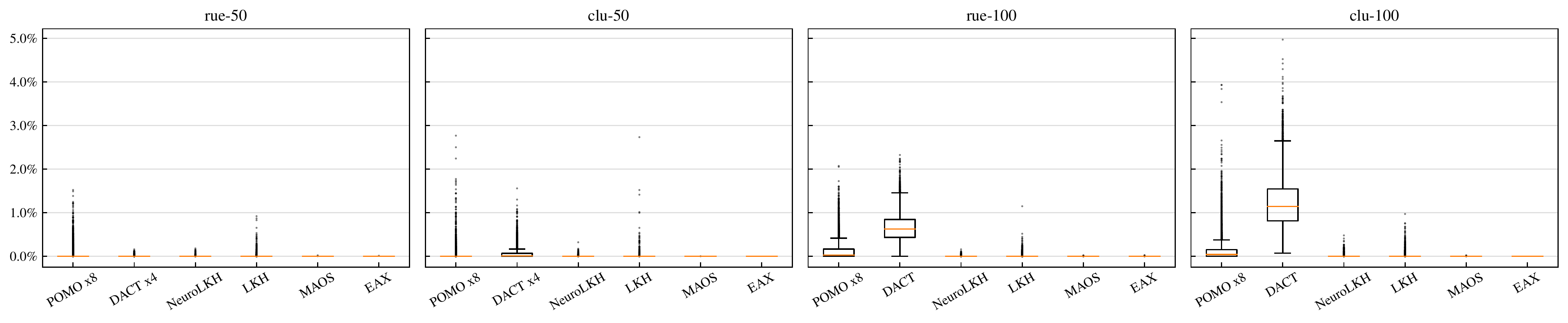}} \\
			& {\large{}TSP solvers}
		\end{NiceTabular}
	}
	\caption{Visual comparison in box plots of the optimum gaps achieved by the tested solvers in Exp\_1.}
	\label{fig:exp_1}
\end{figure*}
\begin{figure*}[tbp]
	\centering
	\resizebox{\textwidth}{!}{
		\begin{NiceTabular}{cc}
			\Block[]{}{\rotatebox{90}{\large{}Optimum gap}} &
			\Block{}{\includegraphics[trim=7pt 10pt 0 0,clip]{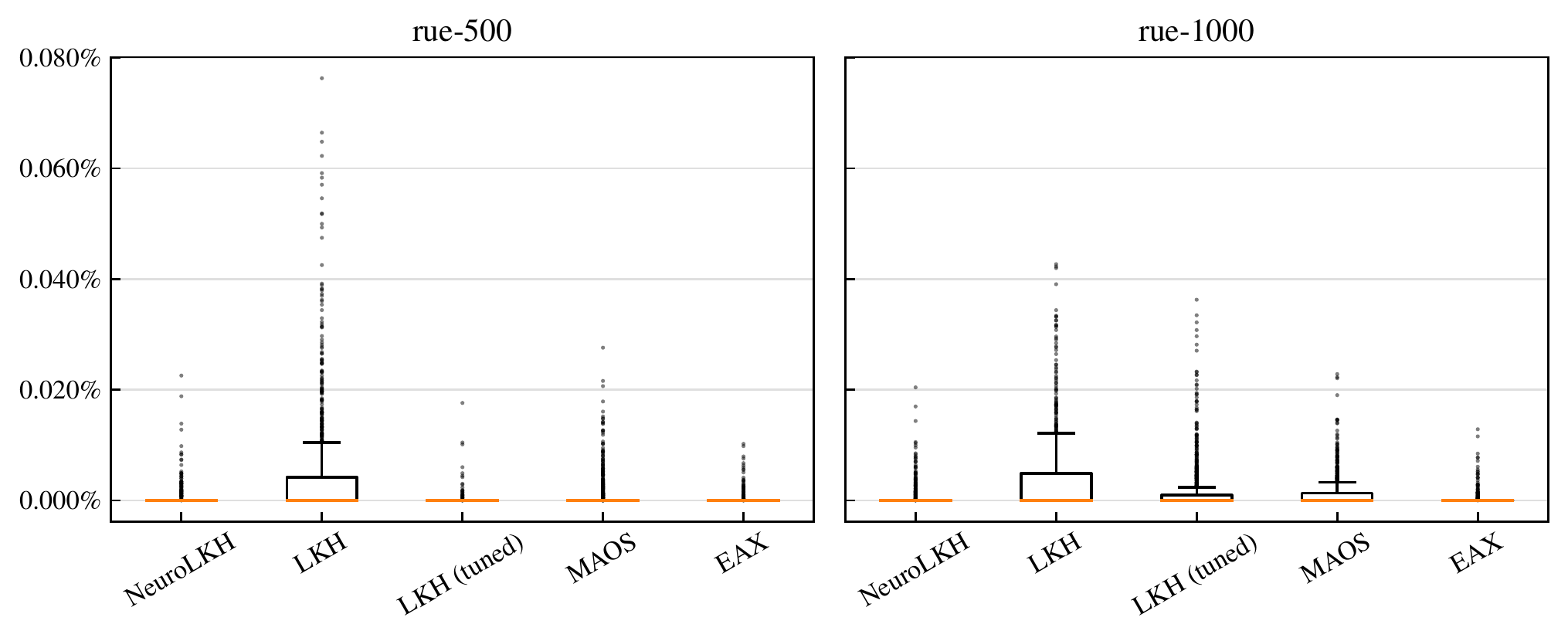}
				\includegraphics[trim=7pt 10pt 0 0,clip]{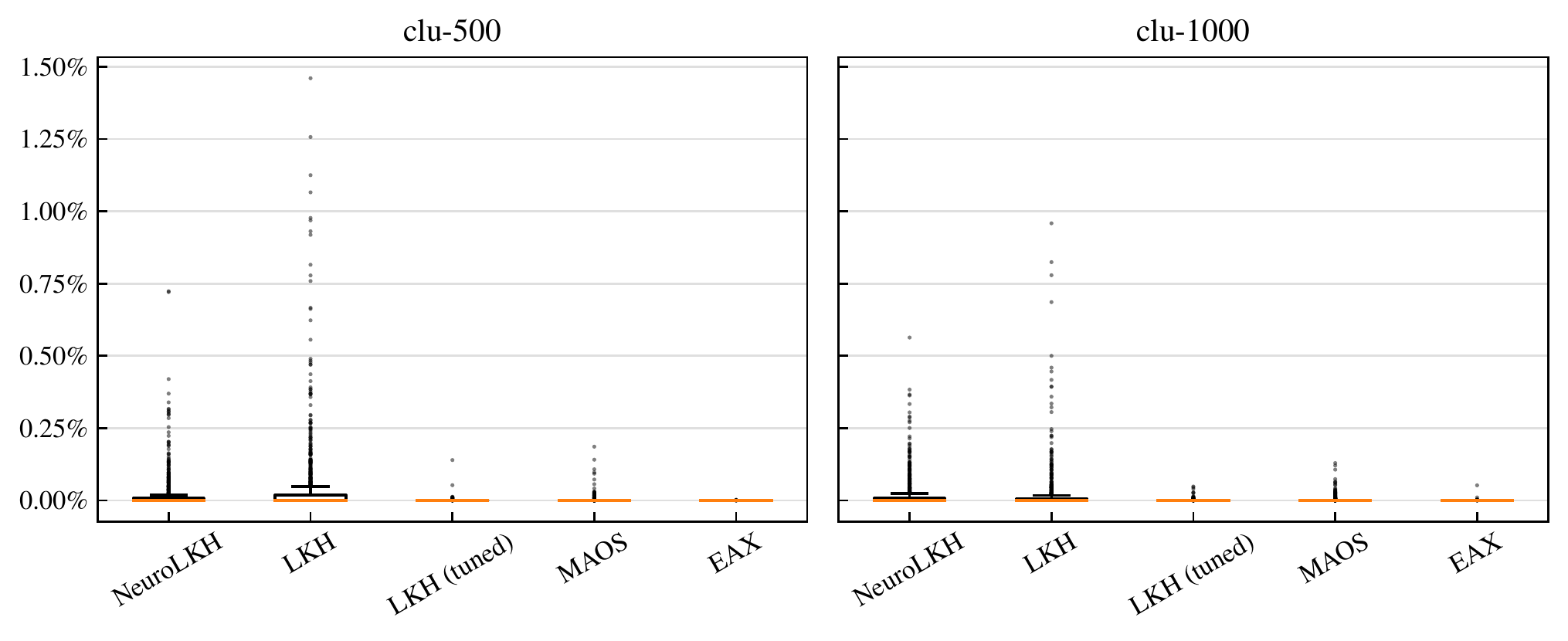}} \\
			&
			\Block{}{\large{}TSP solvers}
		\end{NiceTabular}
	}
	\caption{Visual comparison in box plots of the optimum gaps achieved by the tested solvers in Exp\_2.}
	\label{fig:exp_2}
\end{figure*}

\subsection{Exp\_1: Small-Size Testing Instances with the Same Problem Characteristics as Training Instances}
\label{sec_exp1}
The testing results of Exp\_1 in terms of average optimum gap (Gap),  standard deviation (std), total computation time, and energy are reported in Table~\ref{tab:exp_1}.
In Exp\_1, for DACT with the instance augmentation mechanism (denoted by aug.), its results on the \textit{rue}-100 and \textit{clu}-100 testing instances are missing because it ran for prohibitively long periods of time to solve these instances.
Besides, the tuned variant of LKH, i.e., LKH (tuned), was not included in Exp\_1 because the original LKH has already achieved nearly optimal solution quality.
In addition,  the medians and variance of the optimum gaps across all the testing instances are visualized by box plots in Figure~\ref{fig:exp_1}.
For brevity, the name of the NCO approach is used to denote the solvers learned by it.

The first observation from these results is that traditional solvers still achieved much better solution quality than the learned solvers.
For example, EAX and MAOS could notably solve all the testing instances to optimality.
Among all the randomized solvers, i.e., DACT, NeuroLKH, LKH, EAX, and MAOS, EAX and MAOS also exhibited the best stability.
They achieved the smallest standard deviation over 10 repeated runs.

The second observation is that, after EAX and MAOS, NeuroLKH was the third best-performing solver.
Compared to LKH, NeuroLKH reduced the average optimum gap by one order of magnitude on three out of the four testing sets.
Based on Figure~\ref{fig:exp_1}, one can also observe that NeuroLKH achieved more stable performance than LKH across the testing instances.
Compared to the other two NCO solvers POMO and DACT, the performance advantages of NeuroLKH in terms of solution quality were much more significant.
In general, NeuroLKH could reduce the average optimum gaps by at least two orders of magnitude on all four testing sets. 
Although the performance of POMO and DACT could be improved when equipped with the instance augmentation mechanism, they still performed worse than NeuroLKH.

The third observation is that NCO solvers could generally achieve better solution quality on the \textit{rue} instances than on the \textit{clu} instances.
For example, the average optimum gap achieved by POMO on the \textit{clu}-50 testing instances was 14.18\% greater than that obtained on the \textit{rue}-50 testing instances, and the corresponding numbers for DACT and NeuroLKH were 9.60 times and 33.33\%, respectively.
Moreover, as the problem size grows, such performance gaps became larger.
These results show that current NCO approaches are less adept at learning solvers for structural problem instances (i.e., clustered TSP instances) than for uniformly and  randomly generated instances, indicating that the learning models adopted by them may have limitations in handling structural data.
This could be an important direction for improving NCO approaches.

The fourth observation is that in Figure~\ref{fig:exp_1}, as the problem size grows, the performance of POMO and DACT significantly deteriorated, while the performance of NeuroLKH was still stable.
These results indicate that currently the scalability of LCH approaches and LIH approaches is still quite limited.

The last observation is that regarding efficiency, POMO exhibited excellent performance in terms of both runtime and energy.
Notably, it usually consumed at most one-tenth of the resources consumed by other solvers, which could be very useful in resource-limited environments.
This is also true when EAX and LKH were terminated at the solution quality achieved by POMO solver (marked by LKH* and EAX* in Table~\ref{tab:exp_1}).
On the other hand, considering the poor scalability of POMO, its high efficiency was still limited to small-size problem instances.
It is also found that another NCO approach, DACT, performed poorly in terms of efficiency, especially when equipped with the instance augmentation mechanism.
Finally, NeuroLKH could improve the efficiency of the original LKH in both runtime and energy, and in general, the efficiency of EAX was slightly worse than the LKH-family solvers.

\subsection{Exp\_2: Medium/Large-Size Testing Instances with the Same Problem Characteristics as Training Instances}
\label{sec_exp2}
Similar to Exp\_1, the testing results of Exp\_2 are reported in Table~\ref{tab:exp_1} and illustrated in Figure~\ref{fig:exp_2}.
The main difference between Exp\_1 and Exp\_2 is that the latter considered much larger problem sizes.
In Exp\_2, POMO and DACT were not tested due to their poor scalability.

The first observation from these results is that overall, EAX is still the best-performing solver in terms of solution quality.
Nevertheless, the tuned variant of LKH outperformed EAX on the \textit{rue}-500 testing instances.
Moreover, it outperformed MAOS on three out of the four testing sets, i.e., \textit{rue}/\textit{clu}-500 and \textit{rue}-1000, while the original LKH fell behind MAOS on all four testing sets.
Compared to the original LKH, the tuned variant of LKH could reduce the average optimum gaps by two orders of magnitude on three testing sets and by at least 50\% on the remaining set.
Based on Figure~\ref{fig:exp_2}, it can also be observed that the tuned variant of LKH performed much more stably across the testing instances than LKH.
It is worth mentioning that such performance improvement did not come at the cost of degraded efficiency.
Overall, the tuned variant of LKH and the original LKH performed competitively in terms of time efficiency and energy efficiency. Such results indicate that traditional solvers can largely benefit from parameter tuning and this should be utilized when a sufficient training set is available.

The second observation is that although NeuroLKH could also achieve better solution quality than LKH, the former consumed much more computation time and energy than the latter.
This is particularly evident on the \textit{clu} testing instances.
Such phenomenon once again implies that the current NCO approaches may have limitations on handling structural data.
Taking a closer look at Figure~\ref{fig:exp_2}, one can observe that on the \textit{clu} testing instances, for LKH, parameter tuning could achieve greater performance improvement than NCO (i.e., NeuroLKH).
This may be because parameter tuning can change the behaviors of LKH to a greater extent than that of NeuroLKH (which only modifies the candidate edge set in LKH), eventually leading to better fitting to the specific instance distribution.
Such results also suggest an important future research direction of combining parameter tuning and NCO to achieve more comprehensive control over the behaviors of traditional solvers.


\begin{figure*}[tbp]
	\centering
	\resizebox{\textwidth}{!}{
		\begin{NiceTabular}{cc}
			\Block[]{}{\rotatebox{90}{\large{}Optimum gap}} &
			\Block{}{\includegraphics[trim=8pt 10pt 0 0,clip,height=217pt,valign=t]{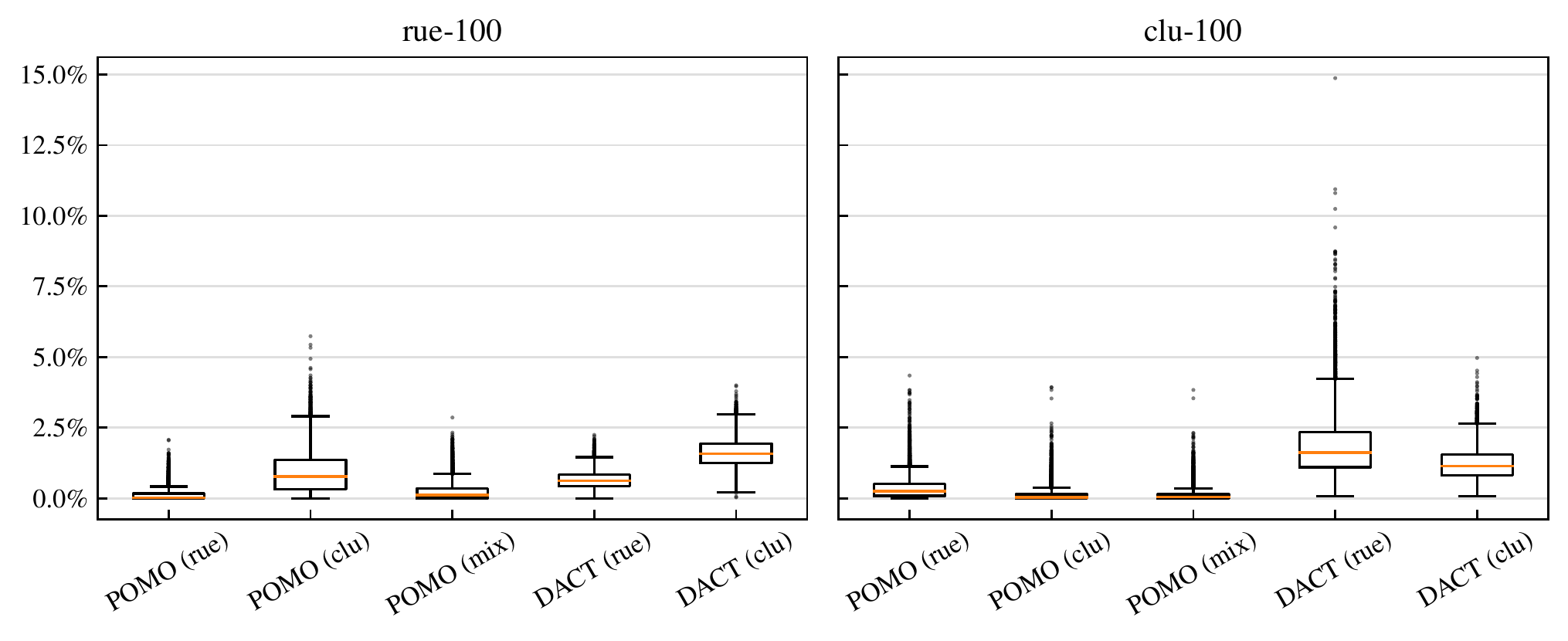}
				\includegraphics[trim=7pt 10pt 0 1pt,clip,valign=t]{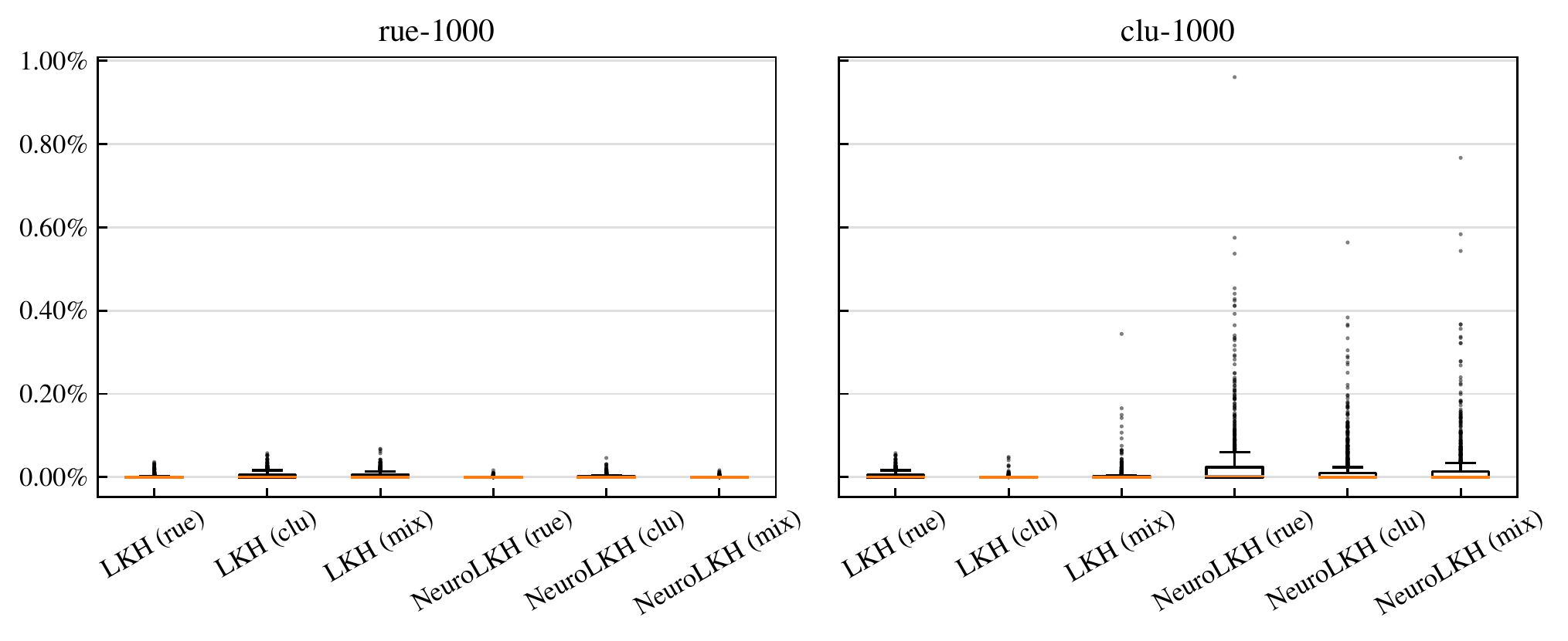}} \\
			&
			\Block{}{\large{}TSP solvers}
		\end{NiceTabular}
	}
	\caption{Visual comparison in box plots of the optimum gaps achieved by the tested solvers in Exp\_3.
	Each learned/tuned solver is marked with the corresponding problem type of the training instances.}
	\label{fig:exp_3}
\end{figure*}

\subsection{Exp\_3: Testing Instances with Different Node Distributions from Training Instances}
\label{sec_exp3}
The medians and variance of the optimum gaps across all the testing instances in Exp\_3 are illustrated by box plots in Figure~\ref{fig:exp_3}.
Note that DACT was unable to converge on the  \textit{mix} training set and thus the corresponding testing performance is not reported.
Recalling that Exp\_3 was designed to assess the generalization ability of the learned/tuned solvers over different problem types, the first observation from Figure 4 is that when applying a learned solver on the testing instances belonging to different types from those of the training instances, the solver's performance degraded.
For example, on the \textit{rue}-100 testing instances, the POMO solver trained on the \textit{rue}-100 training instances performed better than the one trained on the \textit{clu}-100 training instances, and the results were the opposite of the \textit{clu}-100 testing instances.
Based on the last two plots in Figure 4, one can observe that this was also true for NeuroLKH and the tuned variants of LKH.

The second observation is that when a \textit{mix} training set was used, the learned/tuned solver still could not obtain the best possible performance.
For example, on the \textit{rue}-100 testing instances, the POMO solver trained on  the \textit{mix}-100 instances obtained an average optimum gap of 0.2420\%, which is better than the one trained on the \textit{clu}-100 instances (0.9234\%)  but still worse than the one trained on the \textit{rue}-100  instances (0.1278\%).
These results indicate that the adopted learning models may not have sufficient capacity to simultaneously handle different problem types, which could be a direction for improving NCO approaches.

\begin{table}[tbp]
	\centering
	\caption{Testing results of Exp\_4. Each learned solver is marked with the corresponding training set.}
	\label{tab:exp_4}
	\newcommand{\methodcell}{\Block[c]{2-1}{Method\\(training set)}}
	\resizebox{\columnwidth}{!}{
		\begin{NiceTabular}{lc|lc}
			\toprule
			\methodcell{}              & \textit{rue}-100                           & \methodcell{}                & \textit{clu}-100                   \\
			\cmidrule(rl){2-2}
			\cmidrule(rl){4-4}
			& Gap (\%) \(\pm\) std (\%)         &                              & Gap (\%) \(\pm\) std (\%)   \\
			\midrule
			POMO (\textit{rue}-50)              & \hp{0}0.6703 \(\pm\) \hp{0}0.0000 & POMO (\textit{clu}-50)              & \hp{0}0.6829 \(\pm\) 0.0000 \\
			POMO (\textit{rue}-100)             & \hp{0}0.1278 \(\pm\) \hp{0}0.0000 & POMO (\textit{clu}-100)             & \hp{0}0.1405 \(\pm\) 0.0000 \\
			DACT\hspace{2pt} (\textit{rue}-50)  & 27.5437 \(\pm\) 31.4449           & DACT\hspace{2pt} (\textit{clu}-50)  & 21.4630 \(\pm\) 5.3292      \\
			DACT\hspace{2pt} (\textit{rue}-100) & \hp{0}0.6596 \(\pm\) \hp{0}0.5216 & DACT\hspace{2pt} (\textit{clu}-100) & \hp{0}1.2220 \(\pm\) 0.4773 \\
			\bottomrule
		\end{NiceTabular}
	}
\end{table}

\begin{table*}[tbp]
	\centering
		\caption{Testing results of  Exp\_5.
		Each cell contains three values, i.e., number of successes, average optimum gap (\%), and computation time (s). 
		For each testing instance, the highest number of successes is indicated in gray, and the highest number of successes achieved among all LKH variants (including NeuroLKH and the tuned variants of LKH) is boxed.}
	\label{tab:exp_5}
	\newcommand{\partone}[1]{\ifthenelse{#1<10}{\hp{0}}{}#1/10}
	\newcommand{\parttwo}[1]{\ifthenelse{\lengthtest{#1 pt < 10pt}}{\hp{0}}{}#1}
	\newcommand{\partthree}[1]{%
		\ifthenelse{\lengthtest{#1 pt < 10pt}}{\hp{000}}{%
			\ifthenelse{\lengthtest{#1 pt < 100pt}}{\hp{00}}{%
				\ifthenelse{\lengthtest{#1 pt < 1000pt}}{\hp{0}}{}}}#1s}
	\newcommand{\cell}[4][0]{\Block{}{\ifthenelse{#1=1}
			{\fshadow[white][gray!25]{\partone{#2}, \parttwo{#3}}}
			{\ifthenelse{#1=2}
				{\fshadow[black][white]{\partone{#2}, \parttwo{#3}}}
				{\ifthenelse{#1=3}
					{\fshadow[black][gray!25]{\partone{#2}, \parttwo{#3}}}
					{\fshadow[white][white]{\partone{#2}, \parttwo{#3}}}}}\\
			\fshadow[white][white]{\partthree{#4}}}}
	\scalebox{0.86}{
		\begin{NiceTabular}{cccccccccc}
			\toprule
			\Block{1-2}{Problem}           &                              &
			LKH (default)                  & LKH (\textit{rue})                    & LKH (\textit{clu})                  &
			LKH (\textit{mix})                      & NeuroLKH (\textit{rue})               &
			NeuroLKH (\textit{clu})               & NeuroLKH (\textit{mix})               & EAX                            \\
			\midrule
			
			\Block{10-1}{\rotate \textit{TSPlib}}   & d1655                        &
			\cell[3]{10}{0.0000}{7.65}     & \cell{9}{0.0161}{15.18}      & \cell{8}{0.0322}{11.32}      &
			\cell{5}{0.0805}{13.51}        & \cell{6}{0.0006}{106.75}     & \cell{7}{0.0483}{83.18}      &
			\cell[3]{10}{0.0000}{22.51}    & \cell[1]{10}{0.0000}{27.94}                                   \\
			\cmidrule(rl){2-10}
			
			& fl3795                       &
			\cell{0}{81.0510}{3621.86}     & \cell{9}{0.0348}{87.51}      & \cell[3]{10}{0.0000}{10.48}  &
			\cell{6}{1.8073}{89.72}        & \cell{0}{239.956}{3609.97}   & \cell{0}{186.744}{3593.94}   &
			\cell{0}{275.928}{3582.66}     & \cell{1}{15.3622}{3246.70}                                    \\
			\cmidrule(rl){2-10}
			& fnl4461                      &
			\cell{8}{0.0548}{107.55}       & \cell{1}{1.0462}{431.85}     & \cell{0}{2.3882}{341.85}     &
			\cell{3}{0.6135}{248.90}       & \cell[3]{10}{0.0000}{72.61}  & \cell[3]{10}{0.0000}{44.23}  &
			\cell[3]{10}{0.0000}{36.10}    & \cell{8}{0.0438}{1135.55}                                     \\
			\cmidrule(rl){2-10}
			
			& pcb3038                      &
			\cell{6}{0.4575}{110.66}       & \cell{0}{1.4380}{225.28}     & \cell{0}{2.1424}{142.03}     &
			\cell{1}{1.4234}{121.23}       & \cell{9}{0.0363}{110.28}     & \cell[3]{10}{0.0000}{53.85}  &
			\cell{9}{0.0363}{137.30}       & \cell[1]{10}{0.0000}{123.71}                                  \\
			\cmidrule(rl){2-10}
			
			& pla7397                      &
			\cell[3]{10}{0.0000}{598.89}   & \cell{3}{0.5333}{2473.04}    & \cell{9}{0.0389}{956.07}     &
			\cell{5}{1.5073}{1146.88}      & \cell{0}{5.4551}{3438.43}    & \cell{1}{5.1922}{3483.71}    &
			\cell{5}{1.4884}{2229.15}      & \cell{0}{0.3647}{3600.00}                                     \\
			\cmidrule(rl){2-10}
			
			& pr2392                       &
			\cell[3]{10}{0.0000}{2.56}     & \cell{6}{0.0741}{84.41}      & \cell{1}{0.4471}{74.18}      &
			\cell{6}{0.2460}{44.23}        & \cell[3]{10}{0.0000}{4.84}   & \cell[3]{10}{0.0000}{5.27}   &
			\cell[3]{10}{0.0000}{2.46}     & \cell[1]{10}{0.0000}{55.39}                                   \\
			\cmidrule(rl){2-10}
			
			& rl1889                       &
			\cell{0}{2.7675}{161.49}       & \cell{3}{0.1580}{56.52}      & \cell[2]{6}{0.1074}{26.06}   &
			\cell{4}{3.3330}{23.64}        & \cell{1}{3.8037}{147.02}     & \cell{6}{1.5638}{80.23}      &
			\cell{2}{2.4926}{208.98}       & \cell[1]{10}{0.0000}{33.22}                                   \\
			\cmidrule(rl){2-10}
			
			& rl5934                       &
			\cell{1}{2.4027}{615.60}       & \cell{7}{0.0198}{606.66}     & \cell{5}{0.0252}{469.67}     &
			\cell{0}{4.2892}{528.45}       & \cell{5}{1.8883}{654.52}     & \cell{8}{0.9747}{675.75}     &
			\cell[3]{10}{0.0000}{405.83}   & \cell{5}{0.8722}{1939.15}                                     \\
			\cmidrule(rl){2-10}
			
			& u1817                        &
			\cell{1}{6.4160}{130.73}       & \cell{2}{4.7377}{59.46}      & \cell{1}{8.8110}{38.49}      &
			\cell{0}{0.9963}{26.33}        & \cell[2]{2}{4.6503}{210.41}  & \cell{2}{5.5419}{275.88}     &
			\cell{1}{7.5523}{273.12}       & \cell[1]{8}{1.4161}{747.99}                                   \\
			\cmidrule(rl){2-10}
			
			& u2319                        &
			\cell[3]{10}{0.0000}{0.69}     & \cell[3]{10}{0.0000}{0.07}   & \cell[3]{10}{0.0000}{0.05}   &
			\cell[3]{10}{0.0000}{0.03}     & \cell[3]{10}{0.0000}{1.38}   & \cell[3]{10}{0.0000}{1.37}   &
			\cell[3]{10}{0.0000}{1.37}     & \cell{1}{3.0309}{3250.97}                                     \\
			
			\midrule
			
			\Block{10-1}{\rotate \textit{VLSI}}     & icw1483                      &
			\cell[3]{10}{0.0000}{9.39}     & \cell[3]{10}{0.0000}{0.34}   & \cell[3]{10}{0.0000}{0.47}   &
			\cell[3]{10}{0.0000}{3.28}     & \cell[3]{10}{0.0000}{3.59}   & \cell[3]{10}{0.0000}{2.05}   &
			\cell[3]{10}{0.0000}{7.17}     & \cell[1]{10}{0.0000}{23.76}                                   \\
			\cmidrule(rl){2-10}
			
			& dcc1911                      &
			\cell{1}{2.0325}{193.22}       & \cell{3}{1.0944}{69.81}      & \cell{2}{2.0325}{45.91}      &
			\cell{2}{1.2508}{35.51}        & \cell{4}{0.9381}{227.78}     & \cell{1}{1.5635}{311.04}     &
			\cell[2]{7}{0.4690}{112.02}    & \cell[1]{9}{0.1563}{400.82}                                   \\
			\cmidrule(rl){2-10}
			
			& xpr2308                      &
			\cell{4}{1.2467}{141.41}       & \cell{6}{0.8311}{86.09}      & \cell{5}{1.2467}{61.84}      &
			\cell{4}{0.8311}{41.00}        & \cell{9}{0.2770}{151.17}     & \cell{9}{0.2770}{146.09}     &
			\cell[3]{10}{0.0000}{103.27}   & \cell{9}{0.1385}{413.77}                                      \\
			\cmidrule(rl){2-10}
			
			& irw2802                      &
			\cell{3}{1.8996}{198.79}       & \cell{9}{0.3562}{70.05}      & \cell{8}{0.4749}{41.85}      &
			\cell{1}{2.0183}{76.95}        & \cell{8}{0.2374}{164.17}     & \cell[3]{10}{0.0000}{59.08}  &
			\cell[3]{10}{0.0000}{70.69}    & \cell[1]{10}{0.0000}{69.36}                                   \\
			\cmidrule(rl){2-10}
			
			& lta3140                      &
			\cell{8}{0.2102}{221.66}       & \cell{1}{1.7863}{247.91}     & \cell{1}{5.0436}{140.03}     &
			\cell{4}{1.5761}{104.04}       & \cell{7}{0.4203}{417.62}     & \cell[3]{10}{0.0000}{186.98} &
			\cell{9}{0.1051}{167.16}       & \cell[1]{10}{0.0000}{94.62}                                   \\
			\cmidrule(rl){2-10}
			
			& ltb3729                      &
			\cell{4}{1.0151}{785.12}       & \cell{3}{0.8460}{348.86}     & \cell{3}{1.6919}{206.57}     &
			\cell{2}{2.5379}{159.37}       & \cell{2}{1.3535}{716.95}     & \cell{9}{0.0846}{718.67}     &
			\cell[3]{10}{0.0000}{203.96}   & \cell[1]{10}{0.0000}{125.47}                                  \\
			\cmidrule(rl){2-10}
			
			& bgb4355                      &
			\cell{2}{3.4583}{617.08}       & \cell{0}{6.9166}{556.71}     & \cell{1}{8.1742}{297.43}     &
			\cell{1}{6.0520}{210.62}       & \cell{2}{2.5937}{1200.94}    & \cell{2}{3.7727}{1595.97}    &
			\cell[2]{8}{0.9432}{556.52}    & \cell[1]{10}{0.0000}{169.41}                                  \\
			\cmidrule(rl){2-10}
			
			& xqd4966                      &
			\cell{9}{0.0653}{929.77}       & \cell[3]{10}{0.0000}{103.56} & \cell{7}{0.1959}{366.47}     &
			\cell{6}{0.2612}{235.34}       & \cell{0}{9.3262}{3553.00}    & \cell{2}{1.6323}{3113.29}    &
			\cell{1}{1.3058}{3284.33}      & \cell[1]{10}{0.0000}{208.28}                                  \\
			\cmidrule(rl){2-10}
			
			& fea5557                      &
			\cell{6}{0.5827}{490.28}       & \cell{0}{4.8559}{951.59}     & \cell{0}{8.3522}{503.51}     &
			\cell{1}{5.3739}{386.20}       & \cell{0}{4.2085}{1630.28}    & \cell[3]{10}{0.0000}{258.24} &
			\cell{4}{0.7770}{1754.45}      & \cell[1]{10}{0.0000}{225.14}                                  \\
			\cmidrule(rl){2-10}
			
			& xsc6880                      &
			\cell{0}{3.2505}{1831.57}      & \cell{0}{2.5540}{1695.90}    & \cell{0}{5.2008}{952.99}     &
			\cell{0}{5.8509}{740.00}       & \cell{0}{3.1577}{2777.26}    & \cell{0}{2.2754}{3427.23}    &
			\cell[2]{3}{1.2538}{2684.66}   & \cell[1]{7}{0.2322}{1386.64}                                  \\
			
			\midrule
			
			\Block{10-1}{\rotate \textit{National}} & rw1621                       &
			\cell{0}{7.4469}{2193.96}      & \cell[3]{10}{0.0000}{31.75}  & \cell{7}{0.2303}{49.74}      &
			\cell{0}{8.7129}{72.18}        & \cell{3}{0.8829}{2292.70}    & \cell{6}{0.5374}{1331.73}    &
			\cell{2}{1.5354}{1977.19}      & \cell[1]{10}{0.0000}{27.88}                                   \\
			\cmidrule(rl){2-10}
			
			& mu1979                       &
			\cell{7}{0.0691}{125.62}       & \cell[2]{9}{0.0230}{35.92}   & \cell{8}{0.0460}{20.98}      &
			\cell{1}{1.3005}{49.12}        & \cell{0}{9.6463}{1677.92}    & \cell{0}{2.2913}{180.62}     &
			\cell{0}{0.9908}{1724.79}      & \cell[1]{10}{0.0000}{51.38}                                   \\
			\cmidrule(rl){2-10}
			
			& nu3496                       &
			\cell{0}{5.0660}{3583.15}      & \cell{0}{1.2795}{1213.39}    & \cell{0}{4.9827}{360.71}     &
			\cell{0}{6.3871}{340.17}       & \cell{0}{7.6041}{3585.49}    & \cell[2]{3}{1.4147}{3135.37} &
			\cell{3}{1.4771}{2983.63}      & \cell[1]{10}{0.0000}{133.82}                                  \\
			\cmidrule(rl){2-10}
			
			& ca4663                       &
			\cell{0}{0.1372}{500.99}       & \cell{9}{0.0093}{155.47}     & \cell[2]{9}{0.0031}{185.75}  &
			\cell{3}{0.0279}{323.63}       & \cell{0}{1.5483}{470.20}     & \cell{0}{5.8908}{509.78}     &
			\cell{0}{3.5340}{569.64}       & \cell[1]{10}{0.0000}{381.83}                                  \\
			\cmidrule(rl){2-10}
			
			& tz6117                       &
			\cell[2]{2}{0.4510}{3333.54}   & \cell{0}{1.3326}{1322.48}    & \cell{0}{2.2953}{783.38}     &
			\cell{0}{1.8520}{676.98}       & \cell{0}{6.4578}{3552.03}    & \cell{1}{0.2913}{3475.44}    &
			\cell{1}{0.1976}{3490.76}      & \cell[1]{4}{0.1014}{2361.93}                                  \\
			\cmidrule(rl){2-10}
			
			& eg7146                       &
			\cell{0}{9.5392}{2093.50}      & \cell{0}{2.3436}{1611.97}    & \cell{0}{7.1177}{984.30}     &
			\cell{0}{2.0949}{893.78}       & \cell[2]{0}{0.3695}{2053.23} & \cell{0}{6.0735}{2309.80}    &
			\cell{0}{0.6173}{2965.75}      & \cell[1]{9}{0.0174}{1204.65}                                  \\
			\cmidrule(rl){2-10}
			
			& pm8079                       &
			\cell{0}{4.3794}{3471.67}      & \cell{0}{2.1331}{3510.69}    & \cell{0}{2.9777}{2507.03}    &
			\cell{0}{3.2998}{3215.28}      & \cell[3]{0}{1.3700}{3481.58} & \cell{0}{6.1817}{3469.07}    &
			\cell{0}{2.5328}{3474.12}      & \cell{0}{5.0498}{3600.00}                                     \\
			\cmidrule(rl){2-10}
			
			& ei8246                       &
			\cell{3}{0.3056}{1426.69}      & \cell{0}{1.9256}{2152.78}    & \cell{0}{4.4090}{1375.38}    &
			\cell{0}{2.0177}{1344.62}      & \cell{1}{0.5481}{3133.45}    & \cell[3]{8}{0.0728}{1545.12} &
			\cell{8}{0.1164}{1808.24}      & \cell{5}{0.0776}{2418.63}                                     \\
			\cmidrule(rl){2-10}
			
			& ar9152                       &
			\cell{0}{3.6073}{3419.22}      & \cell{0}{6.9052}{3486.53}    & \cell{0}{0.8827}{3108.36}    &
			\cell[2]{0}{0.2701}{3303.10}   & \cell{0}{0.7599}{3416.96}    & \cell{0}{7.5668}{3420.46}    &
			\cell{0}{3.3469}{3416.12}      & \cell[1]{8}{0.0287}{1585.88}                                  \\
			\cmidrule(rl){2-10}
			
			& kz9976                       &
			\cell[2]{2}{0.7741}{2144.65}   & \cell{1}{1.3570}{3304.81}    & \cell{0}{3.0013}{2299.72}    &
			\cell{0}{1.4776}{2223.27}      & \cell{0}{7.9295}{3037.43}    & \cell{0}{1.2817}{3372.06}    &
			\cell{0}{1.4879}{3375.52}      & \cell[1]{9}{0.0047}{1365.65}                                  \\
			
			\bottomrule
		\end{NiceTabular}
	}
\end{table*}

\subsection{Exp\_4: Testing Instances with Different Problem Sizes from Training Instances}
\label{sec_exp4}
Table~\ref{tab:exp_4} presents the testing results of Exp\_4 in terms of average optimum gap.
Recalling that Exp\_4 was designed to assess the generalization ability of the learned solvers over problem sizes, in Table~\ref{tab:exp_4} each learned solver is marked with the corresponding problem sizes of its training instances.

The main observation is that when applying the solvers learned by POMO and DACT to the testing instances with larger sizes than the training instances, the performance of the solvers seriously degraded.
For example, on the \textit{rue}-100 testing instances, the DACT solver trained on the \textit{rue}-50 training instances could only obtain an average optimum gap of 27.53\%, which is generally an unacceptable level of solution quality for TSP.
Such results demonstrate that although the learning models adopted by POMO and DACT can process variable-length inputs, it does not mean that the solvers trained by them can naturally generalize to larger problem sizes. 

\subsection{Exp\_5: Testing Instances with Different  Node Distributions and Problem Sizes from Training Instances}
\label{sec_exp5}
Recall that Exp\_5 was designed to assess the ability of the learned/tuned solvers to generalize from generated instances to real-world instances, where the latter differed from the former in both problem sizes and problem types.
Specifically, only the LKH-family solvers and EAX were tested in Exp\_5 due to the large problem sizes.
Three training sets were used in Exp\_5, i.e., \textit{rue}-1000, \textit{clu}-1000, and \textit{mix}-1000.
Based on each training set, a tuned variant of LKH and a NeuroLKH solver were obtained.
Then, for each testing instance, each solver was applied for 10 runs.
Table~\ref{tab:exp_5} presents the testing results in terms of the number of times where the optimal solution was successfully found among the 10 runs, average optimum gap, and average computation time.

The first observation from Table~\ref{tab:exp_5} is that the best-performing solver is EAX.
It achieved the highest number of successes on 22 out of the 30 testing instances, which is far more than that of the second best-performing solver.
In particular, on the testing instances belonging to the \textit{National} benchmark set, the performance gap between the LKH family solvers and EAX is significant, indicating that LKH may be intrinsically limited in solving this type of instances.

The second observation is that among all the LKH-family solvers, the solver learned by NeuroLKH on the \textit{mix}-1000 training instances succeeded more times than the tuned variants of LKH and the original LKH.
This may be because the mixed training set could cover more cases of possible TSP instances than the pure \textit{rue} or \textit{clu} training sets, finally leading to better generalization.
On the other hand, based on the same training set, the NeuroLKH solver consistently performed better than the tuned variant of LKH.
For example, the variant of LKH tuned on the \textit{mix}-1000 training instances obtained fewer successes than the solver learned by NeuroLKH on the \textit{mix}-1000 training instances, and the former was actually the one among all the LKH-family solvers that achieved the fewest successes.
These results indicate that the parameter tuning process of LKH may cause the solver to be more easily overfitted to the training instances than the NCO approach NeuroLKH.

In summary, for real-world TSP instances that are unknown in advance, EAX is an appropriate default solver to use; to learn/train a TSP solver for these instances, NeuroLKH seems to be a better option than parameter tuning.

\begin{figure}[tbp]
	\centering
	\resizebox{\columnwidth}{!}{
	\begin{NiceTabular}{c}
		\Block{}{\includegraphics{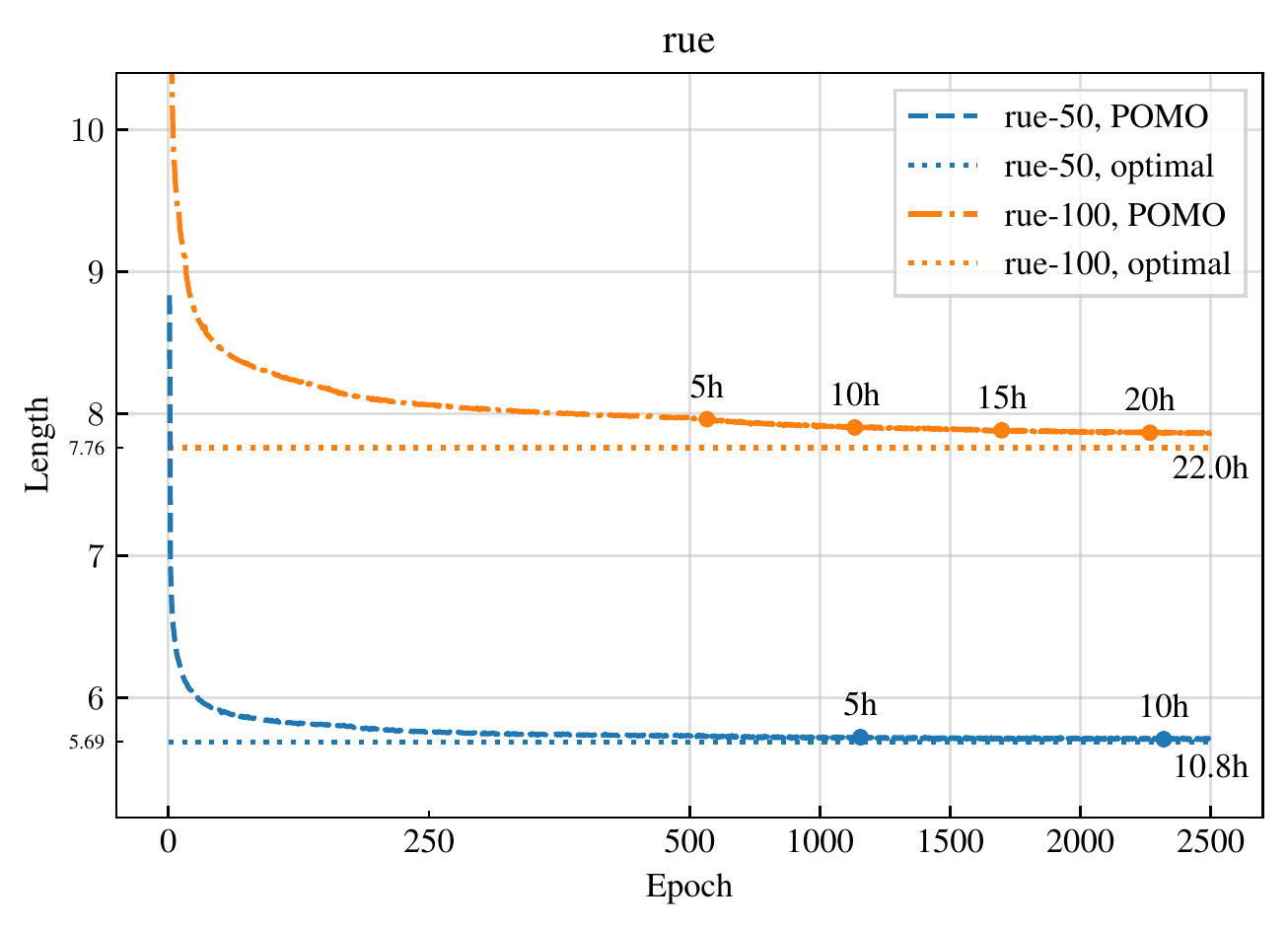}
			\includegraphics{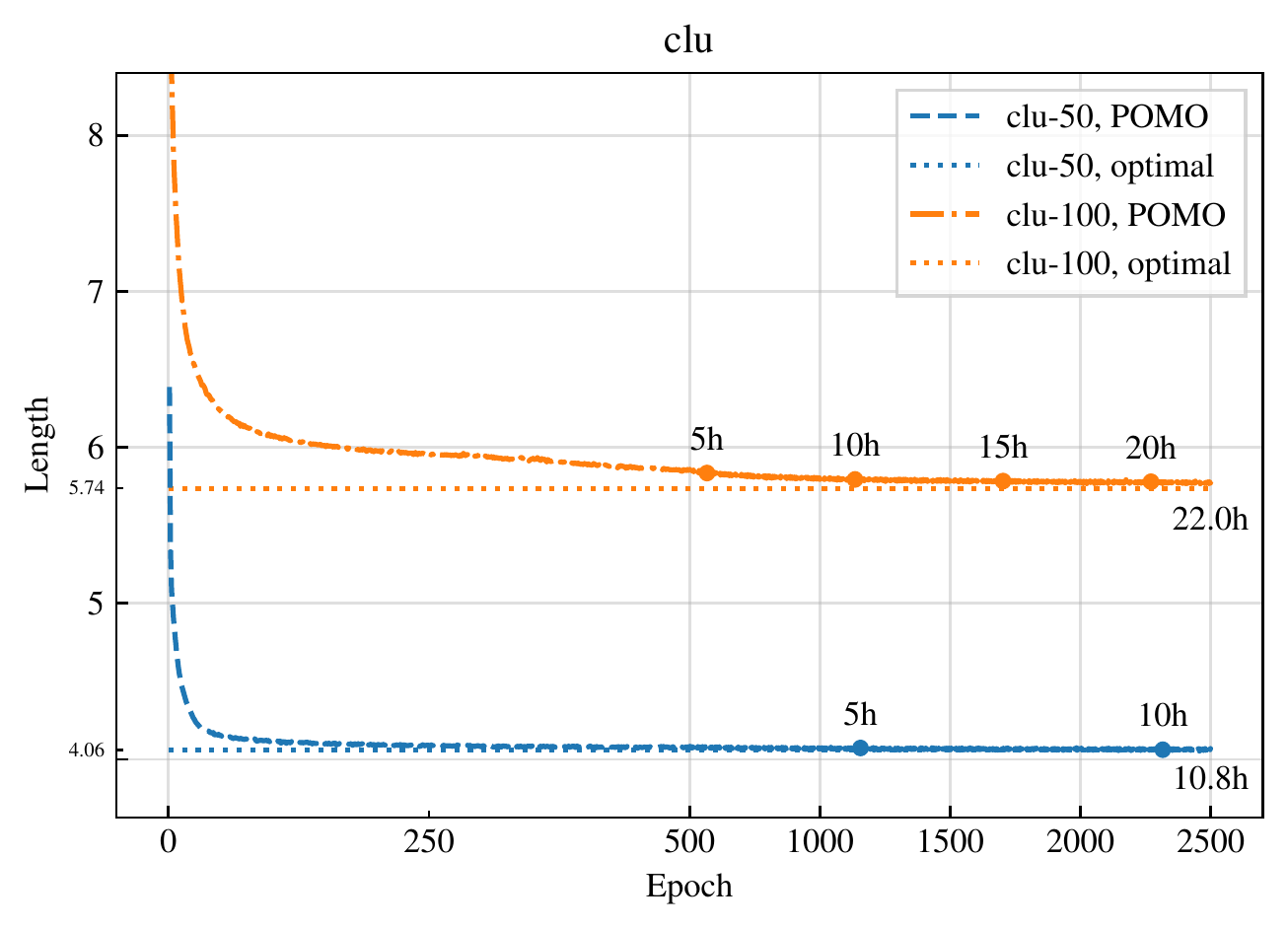}}
	\end{NiceTabular}
}
	\caption{Learning curves of POMO solvers for \textit{rue}-50/100 and \textit{clu}-50/100.}
	\label{fig:learning_curves_POMO}
\end{figure}	

\begin{figure}[tbp]
	\centering
	\resizebox{0.7\columnwidth}{!}{
		\begin{NiceTabular}{c}
			\Block{}{\includegraphics{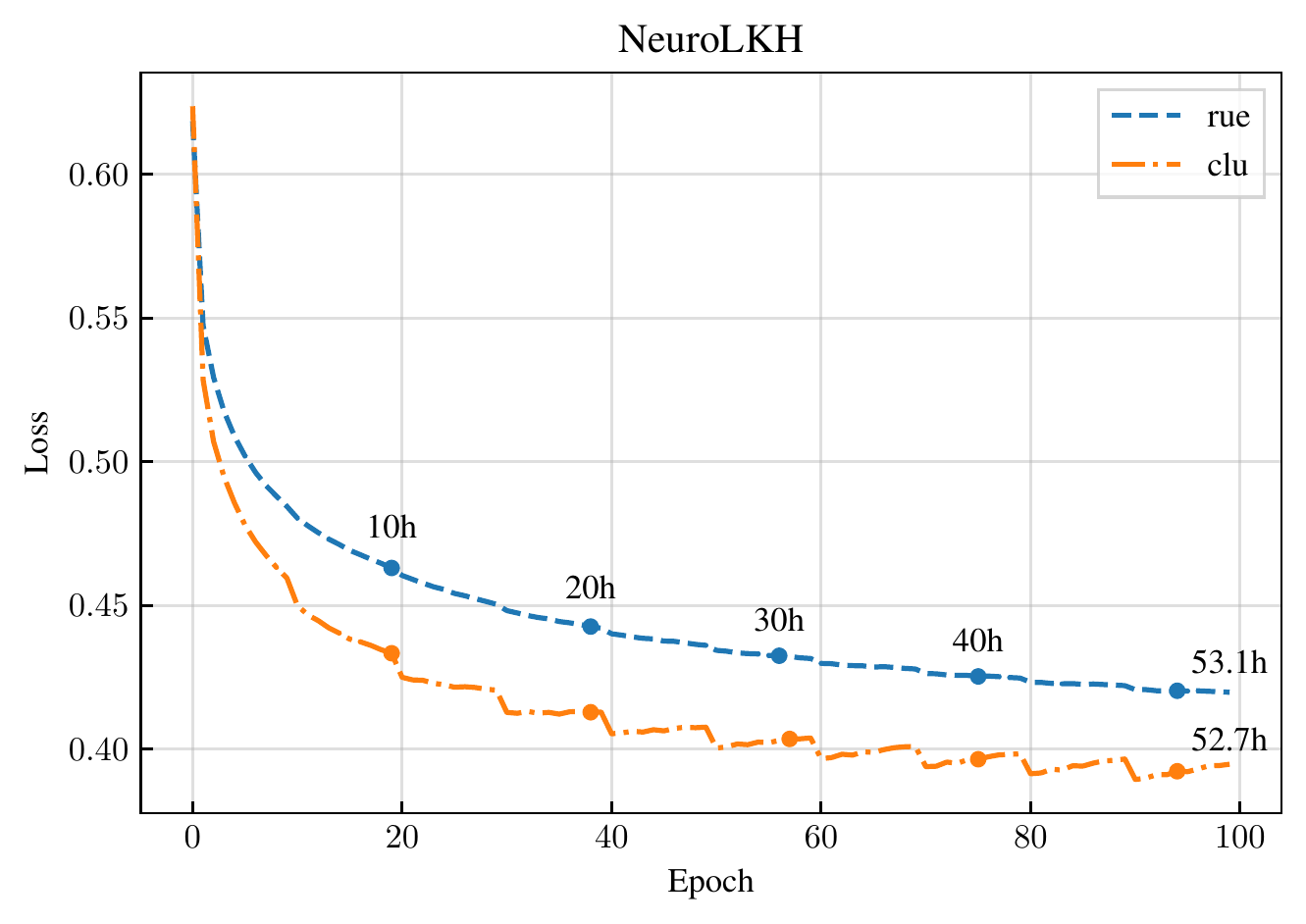}}
		\end{NiceTabular}
	}
	\caption{Learning curves of NeuroLKH for \textit{rue}-500 and \textit{clu}-500 in terms of training loss. The GPU hours consumed for training are also illustrated.}
	\label{fig:learning_curves_nlkh}
\end{figure}	

\subsection{Learning Curves of NCO Solvers}
It is meaningful to investigate the training phases of NCO solvers, since they have a significant impact on the solvers' performance.
Figure~\ref{fig:learning_curves_POMO} illustrates the learning curves of POMO solvers for \textit{rue}/\textit{clu}-50/100.
After each training epoch, the POMO solver was evaluated on a validation set of 10000 problem instances; then, the average tour length of the obtained TSP solutions is plotted. Moreover, the average tour length of the optimal solutions and the GPU hours consumed for training POMO solvers are also illustrated in Figure~\ref{fig:learning_curves_POMO}.
Figure~\ref{fig:learning_curves_nlkh} illustrates the learning curves of NeuroLKH for \textit{rue}/\textit{clu}-500.
Note that for NeuroLKH, the trained model was used to generate a candidate edge set for LKH, not to directly solve the problem instances; thus, in Figure~\ref{fig:learning_curves_nlkh} the training loss is plotted.

From these results, one could make three observations.
First, the validation performance of POMO solvers gradually improved as the training epochs increased and eventually sufficiently converged. 
Second, as the problem size increased from 50 to 100, the learning curves of POMO solvers converged more slowly, and the final optimum gaps became larger.
This echoes the previous finding that the learning capability of POMO solvers is not sufficient for handling large-size problems.
Finally, the training time of NCO solvers could vary from several GPU hours to several GPU days, but it is generally acceptable.

\section{Conclusion}
\label{sec:con}
The applications of neural networks to solve CO problems have been studied for decades (starting from HNN-based works \cite{hopfield1985neural,smith1999neural}), and recently a subfield known as neural combinatorial optimization (NCO) has emerged rapidly.
This work highlighted several issues exhibited by the comparative studies in the existing NCO works and presented an in-depth comparative study of traditional solvers and NCO solvers on TSPs.
An evaluation protocol driven by five research questions was established, which could be used as a basis for benchmarking NCO approaches against others on more CO problems.
Specifically, two practical scenarios, categorized by whether one could collect sufficient training instances to represent the target cases of the problem, were considered.
Then, the performance of the solvers was compared in terms of five critical aspects in these scenarios, i.e., effectiveness, efficiency, stability, scalability, and generalization ability.
Five different problem types with node numbers ranging from 50 to 10000 were used as the benchmark instances in the experiments.

Based on the experimental results, it is found that, in general, NCO solvers were still dominated by traditional solvers in nearly all performance aspects.
A potential benefit of NCO solvers might be their high efficiency (in terms of both time and energy) on small-size problem instances.
It is also found that, for NCO approaches, a crucial assumption is that the training instances should sufficiently represent the target cases of the problem; otherwise, the trained solvers would exhibit severe performance degradation on the testing instances.
However, in many real-world applications, one can only collect a limited number of problem instances \cite{LiuTY22,TangLYY21}, or the accumulated instances are outdated and cannot effectively reflect the current properties of the problem \cite{Reilly09,Smith-MilesB15,TangWLY17}.
In these cases, collecting a good training instance set can take a significant amount of time and may even be impossible, which might reduce the potential advantage of NCO approaches.

As shown in the experiments, NCO faces several challenges that need to be dealt with in the future; several potential research directions are suggested.
\begin{enumerate}
  \item Development of novel architectures or training algorithms to better handle structural problem instances.
  \item Enhancement of current NCO approaches to learn solvers that can perform well on large-size problem instances and multiple (not one) problem types.
  \item Hybridization of parameter tuning and NCO to achieve more comprehensive control over the behaviors of traditional solvers, hopefully leading to even better performance.
\end{enumerate}

Finally, it is worth mentioning that the merits of a CO solver can always be considered from two different perspectives.
The first is its strength, i.e., how well it can solve a particular CO problem, which is exactly the perspective adopted by this work.
The second is the generality, i.e., how many different CO problems it can be used to solve.
Recent studies have extended NCO with unified DNN models to many different CO problems \cite{khalil2017learning,li2018combinatorial}, where there might be no specialized solvers such as LKH and EAX.
Hence, it seems that NCO is a potential alternative for general-purpose CO solvers.
A systematic evaluation study of the generality of NCO approaches has the potential for future research.

\section*{Acknowledgments}
This work was supported in part by the National Key Research and Development Program of China under Grant 2022YFA1004102, in part by the National Natural Science Foundation of China under Grant 62250710682, and in part by the National Natural Science Foundation of China under Grant 62272210.


\bibliography{IEEEabrv,mybib}
%

\bibliographystyle{IEEEtran}

\vfill

\end{document}